%% file: noc.tex
\documentclass[10pt,twocolumn,letterpaper]{article}

\usepackage[accsupp]{axessibility} 

\usepackage{iccv}
\usepackage{times}
\usepackage{epsfig}
\usepackage{graphicx}
\usepackage{amsmath}
\usepackage{amssymb}

\usepackage{enumitem}
\usepackage{booktabs, multirow} 
\usepackage{soul} 
\usepackage[table,dvipsnames]{xcolor} 
\usepackage{caption}
\usepackage{subcaption}
\usepackage{ragged2e}
\usepackage{blindtext}

\usepackage{kotex}

\usepackage[pagebackref=true,breaklinks=true,letterpaper=true,colorlinks,bookmarks=false]{hyperref}

\usepackage[capitalize]{cleveref}
\crefname{section}{Sec.}{Secs.}
\Crefname{section}{Section}{Sections}
\Crefname{table}{Table}{Tables}
\crefname{table}{Tab.}{Tabs.}

\newcolumntype{x}[1]{>{\centering\arraybackslash}p{#1pt}}

\definecolor{light-gray}{gray}{0.50}

\iccvfinalcopy 

\def\iccvPaperID{6017} 

\ificcvfinal\fi

\begin{document}

\title{Noise-aware Learning from Web-crawled Image-Text Data for Image Captioning}

\author{
    Wooyoung Kang\thanks{These authors contributed equally.} \quad Jonghwan Mun$^*$ \quad Sungjun Lee$^*$ \quad Byungseok Roh \\
    Kakao Brain \\
    {\tt\small \{edwin.kang, jason.mun, jun.untitled, peter.roh\}@kakaobrain.com}
}

\maketitle
\ificcvfinal\thispagestyle{empty}\fi

\input{./sec/0_abstract}

\input{./sec/1_introduction}

\input{./sec/2_related_work}

\input{./sec/3_method}

\input{./sec/4_experiment}

\input{./sec/5_conclusion}

{\small
\bibliographystyle{ieee_fullname}
\bibliography{noc}
}

\clearpage
\newpage
\input{./sec/6_appendix}

\end{document}

%% file: sec/0_abstract.tex

\begin{abstract}
Image captioning is one of the straightforward tasks that can take advantage of large-scale web-crawled data which provides rich knowledge about the visual world for a captioning model. However, since web-crawled data contains image-text pairs that are aligned at different levels, the inherent noises (e.g., misaligned pairs) make it difficult to learn a precise captioning model. While the filtering strategy can effectively remove noisy data, it leads to a decrease in learnable knowledge and sometimes brings about a new problem of data deficiency. 
To take the best of both worlds, we propose a \textbf{\underbar{No}}ise-aware \textbf{\underbar{C}}aptioning (NoC) framework, which learns rich knowledge from the whole web-crawled data while being less affected by the noises. This is achieved by the proposed alignment-level-controllable captioner, which is learned using alignment levels of the image-text pairs as a control signal during training. The alignment-level-conditioned training allows the model to generate high-quality captions by simply setting the control signal to the desired alignment level at inference time. 
An in-depth analysis shows the effectiveness of our framework in handling noise.
With two tasks of zero-shot captioning and text-to-image retrieval using generated captions (i.e., self-retrieval), we also demonstrate our model can produce high-quality captions in terms of descriptiveness and distinctiveness. 
The code is available at \url{https://github.com/kakaobrain/noc}.
\end{abstract}


%% file: sec/1_introduction.tex

\section{Introduction}

The recent introduction of large-scale data of image-text pairs~\cite{COYO, LAION, ALIGN} has brought remarkable advances in computer vision, \eg, CLIP~\cite{CLIP} for multi-modal representation learning and DALL$\cdot$E~\cite{dalle} for the text-to-image generation task. 
This is mainly thanks to the scalability of the data collection process as well as the rich knowledge described in alt-texts of web-crawled data.
Inspired by this, research on image captioning is also moving towards exploiting large-scale web-crawled image-text paired data~\cite{simvlm, git, mplug, coca}.

\begin{figure}[t]
    \begin{center}
    \scalebox{0.95}{
        \includegraphics[width=\linewidth]{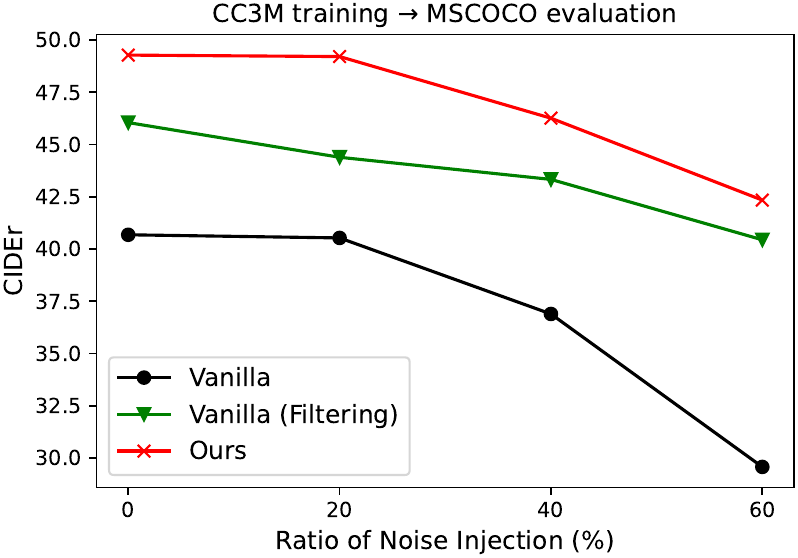}
    }
    \vspace{-0.55cm}
    \end{center}
    \caption[Caption for LOF]{
         \small Zero-shot captioning performance curve on MSCOCO when varying the ratio of noise injection to data of CC3M. To deliberately make noise data, we replace captions up to the specified ratio with ones from randomly selected images in the dataset. Models learned without consideration of noises suffer from performance degradation, even with a data filtering scheme.\footnotemark~In contrast, our model is more robust to noises and provides more accurate captions, indicating the necessity for noise-aware learning.
    }
    \label{fig:poc}
    \vspace{-0.3cm}
\end{figure}

While web-crawled data is effective in learning rich knowledge about the visual world, it inherently suffers from \textit{noise issues} as some text may be unrelated to its paired image.
According to our observation from \cref{fig:poc}, the quality of captions generated by a standard captioning model, when learned without consideration of noises, dramatically deteriorates as more noisy data is included during training.

\footnotetext{For a fair comparison with the filtering scheme reducing the number of training samples, we train all models for the same steps rather than epochs.}

\begin{figure*}[!t]
    \begin{center}
    \scalebox{0.98}{
        \includegraphics[width=\linewidth]{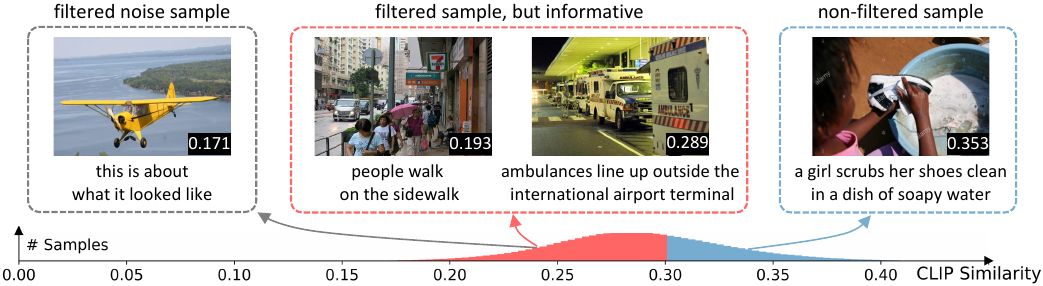}
    }
    \vspace{-0.55cm}
    \end{center}
    \caption{
         \small Examples of web-crawled image-text pairs in CC3M, where numbers within each image indicate CLIP similarity. Filtering with a threshold of 0.3, a selected threshold value on \cite{LAION} after human evaluations, effectively leaves well-aligned samples (right) and removes noise samples (left). However, according to our observation, it often discards informative ones (middle) as well.
    }
    \label{fig:examples}
    \vspace{-0.35cm}
\end{figure*}

One straightforward approach to tackle noises in large-scale web-crawled data is the \textit{CLIP-based filtering strategy}~\cite{LAION} where image-text pairs are filtered out according to their CLIP similarity\footnote{Throughout the paper, the term CLIP similarity is used to denote image-text cosine similarity calculated by the CLIP model, indicating the quality of the caption for a paired image as described in \cite{hessel2021clipscore}.}.
As shown in \cref{fig:poc}, the filtering strategy improves the quality of captions by leaving only relatively well-aligned image-text pairs for training.
However, in general, filtering methods without oracle criteria inevitably discard data informative for training models with CLIP similarity below a certain threshold.
\cref{fig:examples} illustrates such unintentionally filtered cases.
In addition, the inability to access filtered data reduces learnable knowledge, limiting the power of expression during caption generation.
Thus, the filtering strategy may not be optimal for handling noises.

Considering the observation from \cref{fig:examples}, we argue the necessity for the noise-robust image captioning model to fully exploit web-crawled image-text data without filtering;
note that, despite its importance, it is unexplored yet so far as we know.
More specifically, we set our main goal to design a noise-robust model so that the model 1) can generate highly-aligned captions like non-filtered samples in \cref{fig:examples} and 2) also takes advantage of informative knowledge in the data that would be removed with filtering method.

We introduce \textbf{\underbar{No}}ise-aware \textbf{\underbar{C}}aptioning (NoC) framework based on an alignment-level-controllable captioner.
In the framework, we first assign alignment levels to web-crawled data by discretizing CLIP similarities of image-text pairs. 
Then, the model is trained using the alignment level as an additional control signal, enabling the model to generate captions with the desired alignment level.
At inference time, high-quality captions can be generated by feeding a control signal indicating the top level of alignment.

We conduct comprehensive experiments to validate the effectiveness of our model.
First, from the experiments on zero-shot image captioning and self-retrieval tasks, our model outperforms comparative methods, indicating the superior quality of the generated captions in both \textit{descriptiveness} and \textit{distinctiveness}.
Second, we observe that NoC framework enhances the pre-training$\rightarrow$fine-tuning scheme thanks to the more advanced level of visual-language understanding achieved by noise-aware pre-training.
Third, we show that NoC framework provides consistent performance gain compared to the filtering strategy when scaling up dataset sizes up to 125M.
Finally, we further analyze how the noise issue is addressed in the proposed method by investigating the memorization effect of noisy pairs.

Our main contributions are summarized as follows:
\begin{itemize}[label=$\bullet$]
    \vspace{-0.2cm}
    \item 
        We propose a novel Noise-aware Captioning (NoC) framework for handling the noise issue, which is underexplored despite its potential importance. 
        \vspace{-0.2cm}
    \item
        We propose an alignment-level-controllable captioner that utilizes alignment levels of data as a control signal, thus effectively addressing noise issues and being able to generate highly correlated captions by adjusting the control signal at inference time.
        \vspace{-0.2cm}
    \item 
        We show the effectiveness of the proposed noise-aware learning through extensive experiments, where our model outperforms competing methods on both image captioning and self-retrieval tasks with large margins.
\end{itemize}

%% file: sec/2_related_work.tex

\section{Related Work}

\paragraph{Image captioning.}
Various captioning algorithms have been proposed within the encoder-decoder framework~\cite{chen2014learning, vinyals2015show, fang2015captions, mrnn}.
In addition, attention mechanism~\cite{xu2015show, chen2017sca, lu2017knowing, mun2017text} or transformer architecture~\cite{pan2020x, cornia2020meshed, herdade2019image, he2020image, luo2021dual} is incorporated to further boost the performance.
Those models are generally trained on top of human-annotated caption data such as MSCOCO~\cite{lin2014microsoft} and Visual Genome~\cite{krishna2017visual}.
However, learning captioning models from such data has two limitations. 
First, scaling up dataset size is extremely difficult due to expensive human annotations.
Second, the limited learnable knowledge due to small scales results in a poor generalization to the visual concepts in the wild~\cite{agrawal2019nocaps, tran2016rich}.

\begin{figure*}[!t]
    \begin{center}
    \scalebox{0.95}{
        \includegraphics[width=\linewidth]{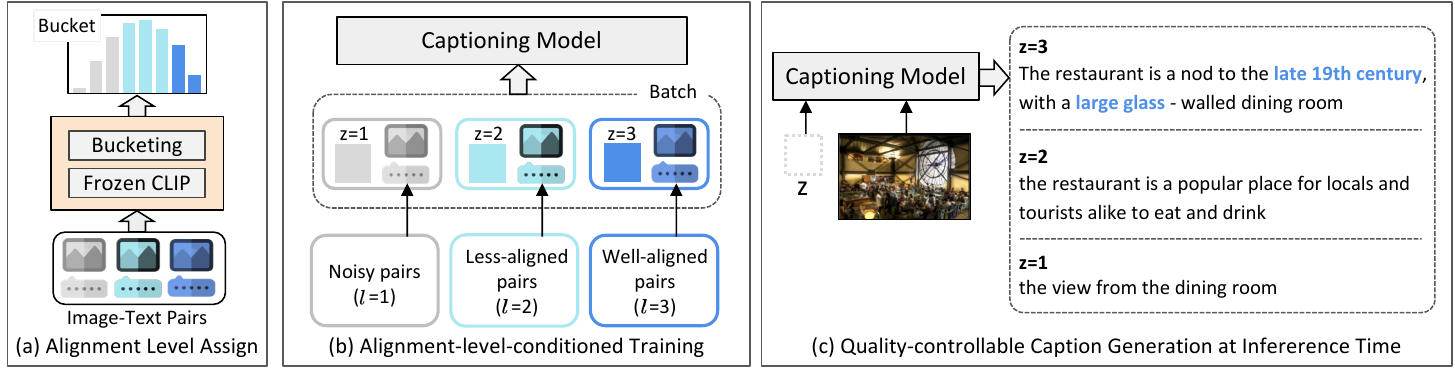}
    }
    \vspace{-0.55cm}
    \end{center}
    \caption{
         \small The proposed noise-aware learning framework from web-crawled image-text data.
         (a) We first identify alignment levels $l$ of individual web-crawled data by discretizing the image-text similarities computed by CLIP.
         (b) Then, we use the alignment levels as the control signal $z$ to train a captioning model (\ie, $z=l$ during training); through the alignment-level-conditioned training, the model is encouraged to generate well-aligned captions with $z=3$ while being guided to generate noisy captions with $z=1$.
         (c) Finally, at inference time, we can generate highly aligned captions by simply feeding a control signal corresponding to the top level of alignment ($z=3$).
    }
    \label{fig:architecture}
    \vspace{-0.35cm}
\end{figure*}

Web-crawled datasets~\cite{cc3m, cc12m, ALIGN, alt200m, LAION} have got attention recently because alt-texts of the data describe paired images with diverse visual concepts, and it is much easy to scale up.
Indeed, some research~\cite{oscar, vinvl, alt200m, simvlm, git, mplug, coca} on image captioning start to exploit the large-scale web-crawled image-text pairs with various vision-language pre-training objectives and show remarkable performances.
However, since the web-crawled data depends on alt-texts, it inevitably includes noisy pairs in a high ratio. 
Although such noisy data may hamper the learning of normal samples, most large-scale research has not yet studied how to effectively handle noisy data; 
existing large-scale learning approaches~\cite{simvlm, coca, alt200m, git} simply train a captioning model without any consideration for noisy pairs.
In contrast, we address the noise issue to maximally exploit all available web-crawled data through noise-aware learning.

\vspace{-0.4cm}
\paragraph{Learning from noisy labels.}
\label{sec:related_work:noisy_labels}
Under the assumption of possible noisy annotations in training data, numerous approaches have been proposed to alleviate the negative effects of the noisy data. Typically, existing works often resort to noise-robust architecture~\cite{adapL, yao2018deep, cheng2020weakly}, specialized design of loss function~\cite{gce, ren2018learning, ma2020normalized} or sample selection~\cite{SELF, han2018co, jiang2018mentornet}. However, most methods consider the unimodal classification task, so it is challenging to extend them to multimodal tasks (\eg, image captioning). 
For example, let us consider one of the most effective approaches, the noise transition matrix~\cite{chen2015webly, adapL}. The noise transition matrix is added on the top of the softmax layer and estimates noisy class posterior probability by discovering the underlying label transition pattern.
However, defining a transition matrix for image captioning is impractical due to the numerous words and long range of contexts. Moreover, other noise-handling algorithms~\cite{han2018co, jiang2018mentornet} that employ additional co-trained networks are inefficient and unsuitable for our large-scale training scenario due to expensive computational costs.

Closely related to our motivation, BLIP~\cite{blip} tackled noisy image-text pairs in web-crawled data with a data bootstrapping technique. However, in addition to web-crawled data, BLIP still relies on supplementary clean human annotations such as MSCOCO to train the captioning and filtering models for bootstrapping.
In contrast, without any clean annotations, we tackle noise issues with the proposed noise-aware learning framework given only web-crawled data.

%% file: sec/3_method.tex

\vspace{-0.25cm}
\section{Noise-aware Learning for Image Captioning}

\subsection{Overview}
\label{sec:method:overview}

Given a pair of an image $I$ and a caption $c$ consisting of $T$ words ($w_1,w_2,...,w_T$), the image captioning models are typically trained by minimizing a negative log-likelihood:
\vspace{-0.3cm}
\begin{equation}
     \mathcal{L} ={-\log p(c|I)} = \sum_{t=0}^{T}{-\log p(w_{t+1}|w_{\le t}, I)},
    \label{eq:log-likelihood}
\end{equation}
where two additional words in \cref{eq:log-likelihood}---$w_0$ ($<$BOS$>$) and $w_{T+1}$ ($<$EOS$>$)---are used to indicate the begin and end of a sentence, respectively. 
In general, the models are designed with the assumption that the training data is clean enough so all image-text pairs are well-aligned.

However, when using the web-crawled data for training, as presented in \cref{fig:poc}, the noisy data hinders the learning of the vanilla captioning models.
A filtering strategy can effectively remove noisy data, but it is also highly likely to discard informative data as depicted in \cref{fig:examples}.
Therefore, to fully benefit from the rich information of web-crawled data, we aim to make a captioning model robust to noise while using the whole dataset.

As illustrated in \cref{fig:architecture}, we propose our NoC framework with an alignment-level-controllable captioner.
During training, we identify an alignment level $l$ of the given image-text pair and use it as a control signal $z$ to make a captioning model be conditioned on the alignment level, which leads to a new objective as follows:
\begin{equation}
    \mathcal{L} = {-\log p(c|I, z)} = \sum_{t=0}^{T}{-\log p(w_{t+1}|w_{\le t}, I, z)}.
\label{eq:loss}
\end{equation}
This objective encourages the model to learn the capability of generating captions of different alignment levels depending on a control signal.
Then, at the inference time, we can steer the captioning model to generate accurate captions by simply feeding a control signal, meaning a top-level of alignment.
With this model design, as discussed in our experiments, we can take the following two advantages: (1) it can learn well-aligned data with less distraction by noisy data and
(2) 
it can still take advantage of data containing useful knowledge (\eg, more diverse visual concepts) that might be discarded when filtering is applied.

In the rest of this section, we first explain how we define alignment levels and assign them to image-text pairs.
Then, we describe the architecture of our alignment-level-controllable captioner.
Finally, we discuss why our method is more effective compared to the filtering strategy.

\subsection{Alignment Level Assignment} 
\label{sec:method:noise_acquisition}

Our key component is how we can assign an alignment level $l$ to a given image-text pair.
For this purpose, we first define the alignment level of a given image-text pair; 
in a nutshell, as an image-text pair is more correlated, we consider the pair is more aligned.
Since there is no ground truth for the correlation, we leverage a pre-trained model (\ie, CLIP) optimized by image-text contrastive learning from web-crawled data;
contrastive learning typically drives a model to learn the correlation between images and texts in a data-driven manner, so the similarity scores by CLIP can be used as alignment scores for image-text pairs.
 
Given the CLIP similarity scores, $s \in \mathbb{R}$, from all training samples, we convert them into $K$ discrete alignment levels $l \in \{1,...,K\}$ via a bucketing technique:
%
\begin{equation}
    l = f_{\text{bucket}}(s),
\end{equation}
where $f_{\text{bucket}}$ is a bucketing function with $K$ bins of equally spaced boundaries.
This bucketing makes more noisy data of low CLIP similarity assigned to a bucket of the lower alignment level (\eg, $l=1$) and well-aligned data of high CLIP similarity allocated to a bucket of the higher alignment level (\eg, $l=K$).

\subsection{Alignment-level-controllable Captioner}
\label{sec:method:caption_generator}
\paragraph{Alignment-level controllability in caption generation.}
Given the image-text data of different alignment levels, our goal is to make a model that generates semantically well-aligned captions (like non-filtered data in \cref{fig:examples}) while benefiting from the whole data.
However, the vanilla captioning model, which processes all data in the same way, is inevitably infected with the noisy samples and is learned to generate captions of limited quality.
Therefore, we introduce a controllable captioning model.
By using alignment levels of image-text data as the control signal (\ie, $z=l$ during training), the quality of captions to be generated can be controllable by adjusting the control signal $z$.
That is, our model is learned to generate well-aligned captions with a control signal of $z=K$ (high alignment level) while generating noisy captions with a control signal of $z=1$ (low alignment level).
At inference time, in practice, we feed a control signal corresponding to a higher alignment level to generate highly well-aligned captions from input images.

\begin{figure}[!t]
    \begin{center}
    \scalebox{1.0}{
        \includegraphics[width=\linewidth]{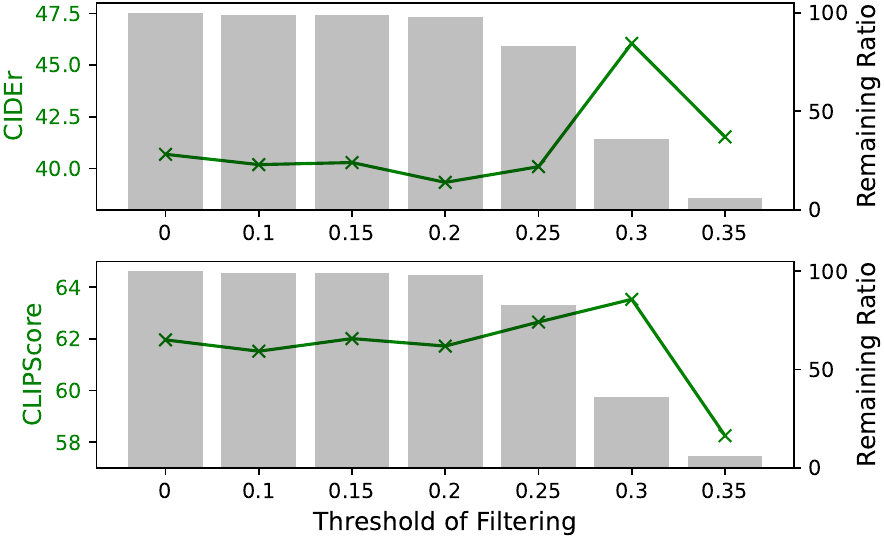}
    }
    \vspace{-0.95cm}
    \end{center}
    \caption{
         \small Captioning scores and the ratio of remaining data when varying the threshold of CLIP similarity score for the filtering. Each model is trained on CC3M and evaluated on MSCOCO. 
    }
    \vspace{-0.1cm}
    \label{fig:filtering_threshold}
\end{figure}

\vspace{-0.3cm}
\paragraph{Architecture.}
Our method is applicable to any captioning model with a simple modification making the decoder take an additional input of the control signal. We employ a VirTex-like~\cite{virtex} transformer-based encoder-decoder model due to its simplicity.
Given a pair of image and text, $(I, c)$, the encoder first extracts a visual feature from the input image.
Next, we feed a control signal $z$---which is an alignment level (\ie, $z=l$) during training calculated as described in \cref{sec:method:noise_acquisition} and is set to a constant one at inference time---into a learnable control embedding layer and then concatenate the resulting control embedding with the image embedding.
Finally, the concatenated vectors are fed into each cross-attention layer of the decoder as key-value, and captions are generated in an auto-regressive manner.
A more detailed explanation is given in \cref{sec:appendix:network_arch}.

\subsection{Discussion}

Recent work~\cite{wang2020diversity} shows that conventional captioning models are trained to generate the so-called \textit{average} captions that consist of common words and phrases in the training corpus.
In other words, with web-crawled data where image and text are aligned at different levels, the captioning models may be trained to generate captions of a common level of alignment (\ie, majority of data), not the highest level. 
With this observation, the filtering strategy can be interpreted as improving the quality of captions by raising the alignment level of common captions.
Accordingly, better performance can be achieved by simply using a higher threshold for filtering.
However, filtering with a higher threshold sometimes leads to a data deficiency problem as it significantly decreases dataset size.
According to our experiment in \cref{fig:filtering_threshold}, the best performance is achieved at a threshold of 0.3, not 0.35 where almost data is discarded.
This implies that the filtering is affected by a trade-off between the quality and the scale of non-filtered data.

In contrast, our NoC framework can address the noise issue in a more principled way.
Controllability allows our model to be thought of as an implicit mixture of experts (but sharing parameters);
one expert (\eg, our model with $z = K$) is specialized to a bucket of highly aligned image-text pairs (like filtering), while parameter sharing between experts allows sharing of learned knowledge for rich visual concepts across all experts (\eg, our model with $z \le K$).

%% file: sec/4_experiment.tex

\section{Experiment}

\subsection{Evaluation Setting}
\label{sec:eval_setting}

Recall that our goal is to learn a captioning model from the web-crawled data, including inherent noises; 
therefore, we evaluate the quality of models without fine-tuning on clean data to check how effectively to tackle noisy data.
To assess the quality of models, we consider two aspects of generated captions: \textit{descriptiveness} (\ie, how well aligned with given images) and \textit{distinctiveness} (\ie, how well describe images with their unique aspects).
To be specific, we conduct zero-shot captioning and self-retrieval tasks (\ie, text-to-image retrieval using generated captions) for \textit{descriptiveness} and \textit{distinctiveness} comparison, respectively. 
Note that the term zero-shot means the model is not further fine-tuned on clean datasets for evaluation.

\vspace{-0.3cm}
\paragraph{Metrics.}
For the zero-shot captioning task, we measure the standard captioning evaluation metrics, \ie, BLEU@4, METEOR, CIDEr, and SPCIE, which compares the generated captions with ground-truth captions.
In addition, we use the CLIPScore~\cite{hessel2021clipscore} to measure whether the generated captions are semantically matched with given images using CLIP.
For the self-retrieval task, we compute the recall (\ie, R@$K$ with $K=\{1,5,10\}$) of the paired images from the generated captions using an external text-to-image retrieval model.
Note that we use the pre-trained CLIP ViT-L/14 for CLIPScore calculation and the retrieval task.

\vspace{-0.3cm}
\paragraph{Datasets.}
When training models, we use Conceptual Captions 3M (CC3M)~\cite{cc3m} dataset.
For the evaluation of zero-shot captioning performance, we exploit MSCOCO~\cite{lin2014microsoft} and nocaps~\cite{agrawal2019nocaps} validation set.
For the self-retrieval task, we leverage MSCOCO and Flickr30k~\cite{plummer2015flickr30k} test set.
Note that, for MSCOCO, we use its Karpathy test split for evaluation.

\subsection{Baselines}
\label{sec:expr_baselines}

Since research on noise-robust image captioning has rarely been explored, we evaluate our method against three carefully designed baselines in a controlled setting to measure their performance without any confounding factors. All baseline models and ours have the same backbone architecture for fair comparisons.

\vspace{-0.3cm}
\paragraph{Vanilla.} 
The first baseline is a vanilla encoder-decoder model and is trained using whole image-text data without any noise handling technique.

\begin{table*}[!t]
    \begin{center}
    \scalebox{0.95}{
    \begin{tabular}{l|cccc|cc|cc|cc|cc}
        \toprule 
        \multirow{3}{*}{Models} &\multicolumn{4}{c|}{\multirow{2}{*}{MSCOCO}} &\multicolumn{8}{c}{nocaps} \\
        & & & & &\multicolumn{2}{c|}{overall} &\multicolumn{2}{c|}{in-domain} &\multicolumn{2}{c|}{near-domain} &\multicolumn{2}{c}{out-of-domain} \\
        &B@4 &M &C &CS &C &CS &C &CS &C &CS &C &CS \\

        \midrule\midrule
        Vanilla &10.31 &15.48 &47.56 &62.89 &41.58 &60.49 &38.60 &58.64 &39.24 &59.91 &51.22 &62.15 \\
        Vanilla (Filtering) &12.81 & 17.30 & 54.66 & 64.84 & 48.96 & 62.70 & 46.06 & 60.74 & 46.33 & 62.35 & 59.50 & 63.92 \\
        Loss weighting & 11.16 & 16.15 & 50.86 & 63.87 & 43.89 & 61.18 & 39.30 & 59.23 & 41.80 & 60.50 & 53.84 & 63.04 \\

        \midrule
        NoC (z=7) & \textbf{15.96} & \textbf{19.50} & \textbf{62.04} & \textbf{66.70} & \textbf{54.94} & \textbf{64.21} & \textbf{51.74} & \textbf{62.54} & \textbf{53.09}  & \textbf{63.92} & \textbf{63.15} & \textbf{65.19} \\
        \bottomrule
    \end{tabular}
    }
    \end{center}
    \vspace{-0.55cm}
    \caption{
        \small Caption generation performance comparison with baselines on MSCOCO and nocaps datasets where all models are trained with CC3M without fine-tuning on the target dataset.
        B@4, M, C, and CS mean BLEU@4, METEOR, CIDEr, and CLIPScore metrics, respectively.
        Numbers in \textbf{bold} indicates the best method.
    }
    \vspace{-0.3cm}
    \label{table:zero_cc}
\end{table*}

\begin{table*}[!t]
    \centering
    \scalebox{0.85}{
        \begin{tabular}{l|ccc|cccc}
            \toprule
            \multirow{2}{*}{Method} & \multirow{2}{*}{Visual Encoder} & \multirow{2}{*}{Text Decoder} & \multirow{2}{*}{Data}  &\multicolumn{4}{c}{MSCOCO} \\
            & & & &BLEU@4 &METEOR &CIDEr &SPICE \\
            \midrule\hline
            \rowcolor[gray]{0.85}\multicolumn{8}{l}{\small \textit{\textbf{Inference time optimization or un-paired training}}} \\ \hline
            ZeroCap~\cite{zerocap} & CLIP ViT-B/32 & GPT2 (345M)~\cite{gpt2} & - & 2.60 & 11.50 & 14.60 & 5.50 \\
            Socratic Models~\cite{socratic} & CLIP ViT-L/14 & GPT3 (175B)~\cite{gpt3} & - & 10.00 & 16.20 & 50.10 & 10.80 \\
            DeCAP~\cite{decap} & CLIP ViT-B/32 & Transformer$_{4\textbf{-}layer}$ (76.5M) & CC3M-text & 8.80 & 16.00 & 42.10 & 10.90 \\
            \midrule\hline
            \rowcolor[gray]{0.85}\multicolumn{8}{l}{\textit{\textbf{\small Supervised training with image-text paired data}}} \\ \hline
            Re-ViLM~\cite{revilm} & CLIP ViT-L/14 & RETRO (410M)~\cite{retro} & CCS + COYO~\cite{COYO} & \textbf{17.90} & - &53.60 & - \\
            SimVLM$_{1.4B}$~\cite{simvlm} & - & - & ALIGN 1.8B~\cite{ALIGN} & 11.20 & 14.70 & 32.20 & 8.50 \\ \hline
            NoC (z=7)$^{\dagger}$ & CLIP ViT-B/32 & Transformer$_{4\textbf{-}layer}$ (76.5M) & CC3M & 14.10 & 18.12 & 48.66 & 12.57 \\
            NoC (z=7)$^{\dagger}$ & CLIP ViT-L/14 & Transformer$_{6\textbf{-}layer}$ (94.5M) & CC3M & 15.96 & \textbf{19.50} & \textbf{62.04} & \textbf{14.37} \\
            \bottomrule
        \end{tabular}
    }
    \vspace{-0.25cm}
    \caption{
        \small Comparison with other models reporting the zero-shot captioning scores. The CCS is a combination of CC3M, CC12M~\cite{cc12m}, and SBU~\cite{sbu} datasets. ${\dagger}$ indicates a method that explicitly handles the problem of noisy samples. Numbers in \textbf{bold} indicate the best method.
    }
    \label{table:sota_comp}
    \vspace{-0.3cm}
\end{table*}

\vspace{-0.3cm}
\paragraph{Vanilla (Filtering).} 
The second baseline is a vanilla captioning model but trained with a filtering strategy to tackle the noise issue.
Following the previous convention~\cite{LAION}, we calculate cosine similarity for each pair of (image, text) using a pre-trained CLIP ViT-B/32 model~\cite{CLIP}, then leave pairs having similarity larger than 0.3 as training data.
The threshold of 0.3 is selected the following \cite{LAION}, which provides the best performance among thresholds from 0.1 to 0.35 with steps of 0.05 as presented in \cref{fig:filtering_threshold}.

\vspace{-0.3cm}
\paragraph{Loss weighting.}
The final baseline is a vanilla captioning model trained with a loss re-weighting strategy to tackle the noise issue.
In this baseline, instead of filtering, we use CLIP similarity score to re-weight sample-wise loss as
\begin{equation}
\label{eq:loss_weight}
    \begin{split}
        \mathcal{L_{\text{weighting}}} = -{1 \over N}\sum_{i=1}^{N}{s_{i}\log p(c_{i}|I_{i})},
    \end{split}
\end{equation}
where $s_i$ indicates a cosine similarity computed by CLIP for $i^{\text{th}}$ image-text pair in a minibatch of $N$ instances. Further details for this baseline are explained in \cref{sec:appendix:loss_weighting}.

\subsection{Implementation details}
We employ a pre-trained CLIP ViT-L/14 for encoding visual features and computing the alignment level $z$, and use 6 randomly-initialized transformer blocks as the caption decoder.
Except for results in~\cref{table:zero_cc,table:sota_comp,table:self_retrieval}, we freeze the visual encoder and take a single CLS token as the visual feature due to its efficiency in all ablation and analytical experiments. More details regarding training settings are described in \cref{sec:appendix:training_details}.
When collecting alignment levels of training samples, we use a bucket of $K$ (=8) bins.
For the data augmentation, we perform the same augmentation strategy of CLIP~\cite{CLIP}, \ie, resizing the shorter side of an original image to 256, then applying a random square crop of 224x224 scale.
We use AdamW optimizer~\cite{adamw} with a linear warm-up strategy for 10\% of whole training iterations followed by learning rate decaying with a cosine schedule. 
We set a base learning rate as 0.0016 with a total batch size of 2048.
During the training phase, we train all baselines and our model for the same iteration steps that correspond to 10 epochs when using the whole data.

\subsection{Zero-shot Captioning Task}
\subsubsection{Comparison with baselines}
\label{sec:comp_with_baselines}
We compare the zero-shot captioning performances of baselines and our method on MSCOCO and nocaps datasets.
In \cref{table:zero_cc}, it has been observed that the Vanilla model exhibits the lowest performance on both the MSCOCO and nocaps datasets, compared to other baselines. This suggests that the absence of any noise-handling technique can impede the training of a model due to the presence of noisy samples.
Also, the filtering strategy provides considerable performance gain compared to the Vanilla and the Loss weighting, which indicates learning a model using well-aligned captions is important for \textit{descriptive} caption generation.
On the other hand, our model significantly outperforms all baselines on both MSCOCO and nocaps datasets. Especially, the notable gain on the nocaps, which covers more diverse visual concepts, implies the importance of noise-aware learning across the entire dataset for acquiring a broader range of visual knowledge compared to learning from filtered data.

\subsubsection{Comparison with other concurrent works}
While the primary goal of our experiments is to measure the noise-robustness, we provide additional comparisons with other works that report zero-shot performance on the MSCOCO to better contextualize the effectiveness of our work in~\cref{table:sota_comp}.
The reported works are divided into two groups: 1) captioning with inference time optimization~\cite{zerocap,socratic} or leveraging an unpaired training strategy~\cite{decap}, and 2) directly training entire networks~\cite{simvlm} or only intermediate modulation network~\cite{revilm} with image-text pairs, but not MSCOCO. 
Due to differences in architecture, training strategy, and dataset, we also provide comprehensive information about the settings for each method.
As shown in \cref{table:sota_comp}, except for the BLEU@4, our algorithm outperforms others in all metrics with a substantial margin, despite having significantly fewer parameters.
Specifically, Re-ViLM~\cite{revilm} shows a better BLEU@4 score than our model. We conjecture that Re-VILM would benefit from the n-grams in retrieved captions by their retrieval augmentation technique at inference time. While the much higher CIDEr score of our model compared to Re-ViLM indicates NoC generates captions with more diverse expressions considering the TF-IDF weighting of the CIDEr. 




\begin{table}[!t]
    \begin{center}
    \scalebox{0.75}{
    \begin{tabular}{l|rrr|rrr}
        \toprule
        \multirow{2}{*}{Models} &\multicolumn{3}{c|}{MSCOCO} &\multicolumn{3}{c}{Flickr30k} \\
        &R@1 &R@5 &R@10 &R@1 &R@5 &R@10 \\

        \midrule
        \textcolor{light-gray}{GT Caption} &\textcolor{light-gray}{34.57} &\textcolor{light-gray}{59.30} &\textcolor{light-gray}{69.91} &\textcolor{light-gray}{63.08} &\textcolor{light-gray}{86.50} &\textcolor{light-gray}{92.00} \\
        
        \midrule
        Vanilla & 25.44 & 50.38 & 61.66 & 47.10 & 76.60 & 85.90 \\
        Vanilla (Filtering) & 31.64 & 58.90 & 70.36 & 56.50 & 85.50 & 92.50 \\
        Loss weighting & 28.78 & 54.44 & 65.44 & 48.00 & 78.90 & 87.50 \\

        \midrule
        NoC (z=7) & \textbf{40.00} & \textbf{66.78} & \textbf{77.53} & \textbf{65.10} & \textbf{92.00} & \textbf{96.20} \\
        \bottomrule
    \end{tabular}
    }
    \end{center}
    \vspace{-0.55cm}
    \caption{
        \small Comparison of self-retrieval capability on MSCOCO and Flickr30k datasets.
        Numbers in \textbf{bold} indicate the best method.
    }
    \vspace{-0.35cm}
    \label{table:self_retrieval}
\end{table}

\subsection{Self-retrieval Task}
We compare self-retrieval capability to measure how well each algorithm \textit{distinctively} describes given images.
\cref{table:self_retrieval} presents a comparison of self-retrieval performances between baselines and our method on MSCOCO and Flickr30k datasets.
From the~\cref{table:self_retrieval}, our method outperforms all three baselines with large margins.
One interesting observation is that generated captions by our method show a higher performance compared to ground-truth captions. We conjecture that the ground-truth captions are semantically accurate but may lack distinctiveness because human annotators would not be explicitly instructed to describe images in a way that they are distinguishable from others.
In contrast, our model can generate highly aligned captions with fine-grained details for the given images by adjusting the control signal with a high alignment level.
This result implies our model is effective in generating \textit{distinct} captions.
The qualitative examples of the retrieval results are illustrated in \cref{fig:self_retrieval_examples} and \cref{sec:appendix:more_qualitative_results}.


\subsection{In-depth Analysis}
\subsubsection{Is it effective after fine-tuning?}
\label{sec:finetuning}
\begin{figure}[t]
    \begin{center}
    \scalebox{0.99}{
        \includegraphics[width=\linewidth]{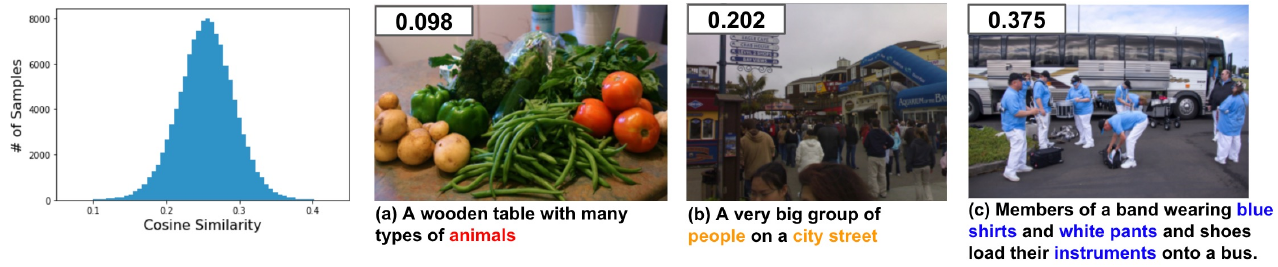}
    }
    \end{center}
    \vspace{-0.50cm}
    \caption{
         \small Distribution of CLIP scores on MSCOCO dataset and examples at different alignment scores. Best viewed in zoom-in.
    }
    \label{fig:coco_example_1}
\end{figure}

We present results after fine-tuning CC3M pre-trained models on MSCOCO in \cref{tab:finetuning}.
It turns out that our NoC framework can enhance the pre-training$\rightarrow$fine-tuning scheme, which is the most common training pipeline for image captioning.
We conjecture that there are two reasons for the effectiveness of our method compared to the filtering baseline as a pre-training method:
1) strong noise robustness in our method enables pre-trained models to acquire a more advanced level of visual-language understanding,
2) as even human-annotated data (\eg, MSCOCO) is aligned at different levels in \cref{fig:coco_example_1}, our NoC framework can be favorably and effectively adapted to fine-tuning with the human-annotated data and leads to performance gains.
It is noticeable that this experimental evidence emphasizes the practical usefulness and potential of our method for delivering improved results even in scenarios where human-annotated data is included.

\begin{table}[t]
    \centering
    \caption{
        Performances on MSCOCO and nocaps after fine-tuning on MSCOCO.
    }
    \vspace{-0.25cm}
    \scalebox{0.90}{
        \begin{tabular}{l|cc|cc}
            \toprule
            \multirow{2}{*}{Method} & \multicolumn{2}{c|}{MSCOCO test split} & \multicolumn{2}{c}{nocaps (overall)} \\
            & CIDEr & SPICE & CIDEr & SPICE \\
            \hline\hline
            Vanilla (Filtering) & 126.14 & 22.37 & 87.03 & 12.52 \\
            NoC      & \textbf{129.09} & \textbf{23.23} &  \textbf{93.33} &  \textbf{13.40} \\ 
            \bottomrule
        \end{tabular}
    }
    \label{tab:finetuning}
    \vspace{-0.15cm}
\end{table}

\vspace{-0.2cm}
\subsubsection{Is it effective when scaling-up data size?}
\label{sec:coyo}
\vspace{-0.1cm}

To validate whether our model is effective in larger-scale web-crawled data, we leverage COYO~\cite{COYO} dataset, which consists of 700M web-crawled image-text data without a CLIP-based filtering scheme.
From COYO, we create four datasets of 3M, 10M, 23M, and 125M scales where the smaller dataset is a randomly sampled subset of the larger ones.
For 10M, 23M, and 125M scales, we use a larger base learning rate (\ie, 0.0032) and a mini-batch size (\ie, 8192), respectively, to train models.
Using four datasets, we train two baselines---Vanilla and Vanilla (Filtering)---and our method;
due to its limited effectiveness compared to Vanilla (Filtering), we do not compare Loss weighting.

\begin{figure}[!t]
    \begin{center}
    \scalebox{0.95}{
        \includegraphics[width=\linewidth]{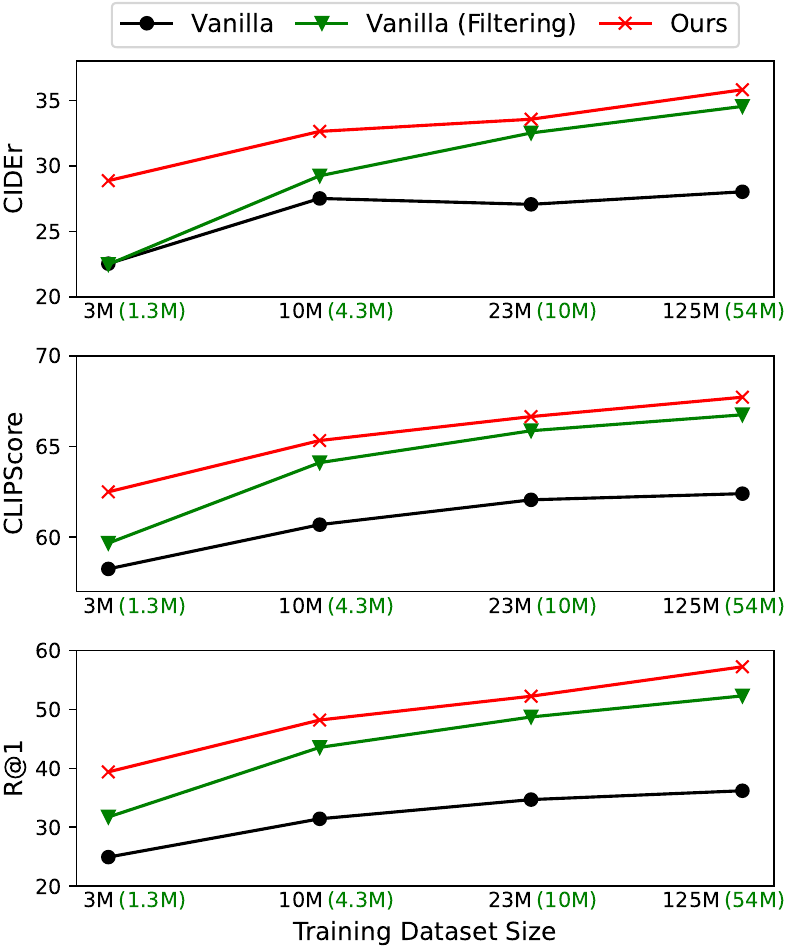}
    }
    \vspace{-0.6cm}
    \end{center}
    \caption{
         \small Zero-shot caption generation (CIDEr and CLIPScore) and self-retrieval (R@1) performance on MSCOCO when scaling up the training dataset sizes using COYO. 
         Note that the green-colored numbers in parentheses mean dataset size after filtering.
         Models of each dataset are trained for the same number of steps.
    }
    \label{fig:coyo_scale_up}
\end{figure}

\cref{fig:coyo_scale_up} summarizes the results where we observe the followings. 
First, due to noisy data, the performance of Vanilla model is quickly saturated compared to others, thus showing a larger gap in the 125M dataset compared to the 3M one.
Second, while Vanilla (Filtering) is considerably enhanced by using larger datasets, the performance gap is consistently kept across larger-scale datasets (23M $\rightarrow$ 125M).
These observations indicate our method is more effective than two baselines in large-scale data as well.

\begin{figure*}[!t]
    \begin{minipage}[b]{0.635\textwidth}
        \begin{center}
        \scalebox{0.92}{
            \begin{tabular}{@{}l|cccc|cccc@{}}
                \toprule
                \multirow{2}{*}{Method} & \multicolumn{4}{c|}{Noisy Group} & \multicolumn{4}{c}{Well-aligned Group} \\ 
                & EM    & B@4   & CIDEr  & CS  & EM     & B@4     & CIDEr   & CS  \\ \midrule
                Vanilla     & 5350    & 12.63  & 122.24  & 52.49  & 8092    & 43.46  & 407.10  & 82.67  \\
                NoC (z=3) & 10527   & 20.41   & 210.83  & 42.87  & 225     & 7.43   & 53.33   & 56.35  \\
                NoC (z=5) & 939     & 5.83   & 48.47   & 55.86  & 4882    & 33.32  & 304.18  & 77.69  \\
                NoC (z=7) & 47      & 2.10  & 18.28   & 59.57  & 10562   & 52.78  & 504.49  & 86.57  \\ \bottomrule
            \end{tabular}
        }
        \end{center}
        \vspace{-0.5cm}
        \captionof{table}{\small Comparison of memorization capability on CC3M training samples of two different noise levels. 
        Note that higher EM (\# Exact Matching), B@4 (BLEU@4), and CIDEr scores in the noisy group mean over-memorization to noisy data. Higher CS (CLIPScore) means better alignment with given images.
        }
        \label{tab:memorization}
    \end{minipage}
    \hfill
    \begin{minipage}[b]{0.34\textwidth}
        \begin{center}
        \scalebox{0.95}{
        \includegraphics[width=\linewidth]{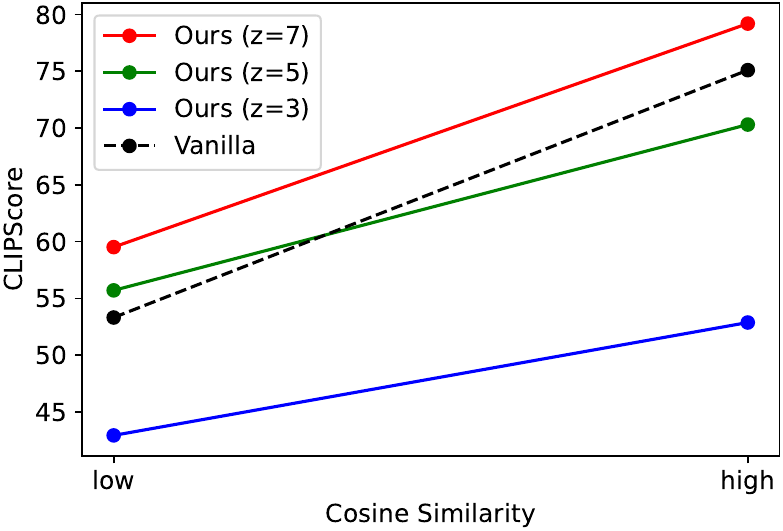}
        }
        \vspace{-0.7cm}
        \end{center}
        \captionof{figure}{\small Performances on two groups of different similarities from CC3M validation set.}
        \label{fig:cc3m_val}
    \end{minipage}
    \vspace{0.2cm}
\end{figure*}

\begin{table*}[!t]
    \centering
    \begin{minipage}[b]{0.33\linewidth}
        \begin{center}	
        \scalebox{0.98}{
            \begin{tabular}{l|cc}
                \toprule
                Options & CIDEr & CLIPScore \\
                \hline \hline
                Quantile & 49.22 & 65.72 \\
                \textbf{Uniform}  & 49.18 & 66.65 \\
                \bottomrule
            \end{tabular}
            \label{tab:ablation:binning}
        }
        \vspace{-0.3cm}
        \caption*{\small (a) Bucketing strategy}
        \end{center}
    \end{minipage}
    \begin{minipage}[b]{0.33\linewidth}
        \begin{center}	
        \scalebox{0.98}{
            \begin{tabular}{l|cc}
                \toprule
                Options & CIDEr & CLIPScore \\
                \hline \hline
                Sum     & 48.99 & 65.19 \\
                MLP     & 48.70 & 66.94 \\
                \textbf{Concat.} & 49.18 & 66.65 \\
                \bottomrule
            \end{tabular}
            \label{tab:ablation:fusion}
        }
        \vspace{-0.3cm}
        \caption*{\small (b) Control fusion method}
        \end{center}
    \end{minipage}
    \begin{minipage}[b]{0.33\linewidth}
        \begin{center}	
        \scalebox{0.98}{
            \begin{tabular}{c|cc}
                \toprule
                Options & CIDEr & CLIPScore \\
                \hline \hline
                4  & 49.28 & 65.67 \\
                \textbf{8}  & 49.18 & 66.65 \\
                16 & 48.58 & 65.67 \\
                \bottomrule
            \end{tabular}
            \label{tab:ablation:num_bin}
        }
        \vspace{-0.3cm}
        \caption*{\small (c) The number of bucket bins}
        \end{center}
    \end{minipage}
    \vspace{-0.65cm}
    \caption{
        \small Ablations on the zero-shot MSCOCO captioning after training on the CC3M dataset.
        The options, highlighted in bold, are selected as our default model due to their balanced performance considering both CIDEr and CLIPScore.
    }
    \label{tab:ablation}
\end{table*}

\vspace{-0.1cm}
\subsubsection{How can our model handle noises?}
\vspace{-0.1cm}
We examine why our algorithm is robust to noisy data by inspecting the captioning results in different alignment levels.
One of the reasons for the degradation of generalization performance when using noisy datasets is the powerful memorization ability of modern DNNs~\cite{memorization, memorization2};
this results in the over-memorization of noisy data in the training set, hindering the learning of normal data.

We analyze such memorization capability of Vanilla and our models for image-text data of different alignment levels.
For this experiment, we train models for longer steps so that the models fully fit and memorize the training data (CC3M).
Then, we compare the memorization capability for \textit{training samples split} into two groups with different alignment levels: 1) a noisy group of CLIP similarity between 0 and 0.15, and 2) a well-aligned group of CLIP similarity higher than 0.35. 
To measure the degree of memorization, we count the number of exact matching (EM) samples, where the generated caption is identical to the paired web-crawled caption of the input image, in addition to captioning metrics.

From \cref{tab:memorization}, our model with $z=7$ seems to be trained mainly from the data of high similarity; thus showing a larger number of exact matching examples in the group of high similarity, but an extremely small number of exact matching cases in the group of low similarity. In contrast, our model with $z=3$ shows the opposite behavior.
Based on this observation, we analyze that our controllable model is implicitly specialized in the different groups according to a control signal ($z$).
As a result, our model with $z=7$ can be less affected by noisy data that is dealt with $z \leq 3$.
On the other hand, the Vanilla baseline suffers from over-memorization to noisy data, thus hindering the learning for pairs of high similarity.
Consequently, by setting $z$ to 7 at inference time, our model outperforms the Vanilla baseline on the CC3M validation set as presented in \cref{fig:cc3m_val}.

\vspace{-0.25cm}
\subsubsection{Ablation study}
We also analyze the impact of two options for model design choice---a binning scheme for bucketing and a fusion method for image and control embeddings---and a hyper-parameter, the number of buckets for noise levels $K$.

\paragraph{Bucketing strategy.}
We compare different discretization strategies for constructing bins of bucketing: 1) \textit{Uniform} bucketing where all bins have identical intervals of CLIP similarity and 2) \textit{Quantile} bucketing where each bin has adaptive widths for containing an equal number of samples.
\cref{tab:ablation}(a) presents results where the two strategies show similar CIDEr scores. Uniform bucketing achieves a higher CLIPScore than Quantile bucketing. 

\vspace{-0.4cm}
\paragraph{Control fusion method.}
We compare different operations for fusing the control and image embeddings to make an input embedding for the decoder: 1) element-wise summation,  2) concatenation in sequence direction, and 3) concatenation in channel dimension followed by MLP.
\cref{tab:ablation}(b) shows that concatenation brings the best-balanced performance when considering both CIDEr and CLIPScore.
We conjecture that this is partly because the condition information remains using the concatenation operation in the fused embedding and thus is more appropriate to directly control the caption generation, while the condition information is smoothed with visual ones in other operations.

\vspace{-0.4cm}
\paragraph{The number of bucket bins.}
\cref{tab:ablation}(c) shows the results across the number of bucket bins $K = \{4, 8, 16\}$.
While our method seems robust to the number of bucket bins, the lowest performance with higher $K$ implies too fine-level bucketing for alignment levels may slightly hinder the learning.

\begin{figure*}[!t]
    \begin{center}
    \scalebox{0.98}{
        \includegraphics[width=\linewidth]{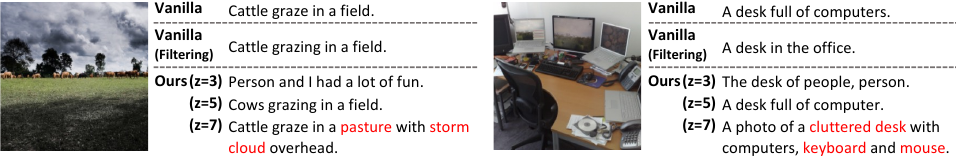}
    }
    \vspace{-0.55cm}
    \end{center}
    \caption{
         \small Examples of generated captions. 
         Compared to the baselines, our model can generate captions of different quality by adjusting a control signal ($z$);
         as we feed $z$ meaning higher alignment levels ($3 \rightarrow 5 \rightarrow 7$), captions become more descriptive and distinct with expressions (highlighted in red) capturing fine details from images.
    }
    \label{fig:qualitative}
\end{figure*}

\begin{figure*}[!t]
    \centering
    \scalebox{0.98}{
        \includegraphics[width=\linewidth]{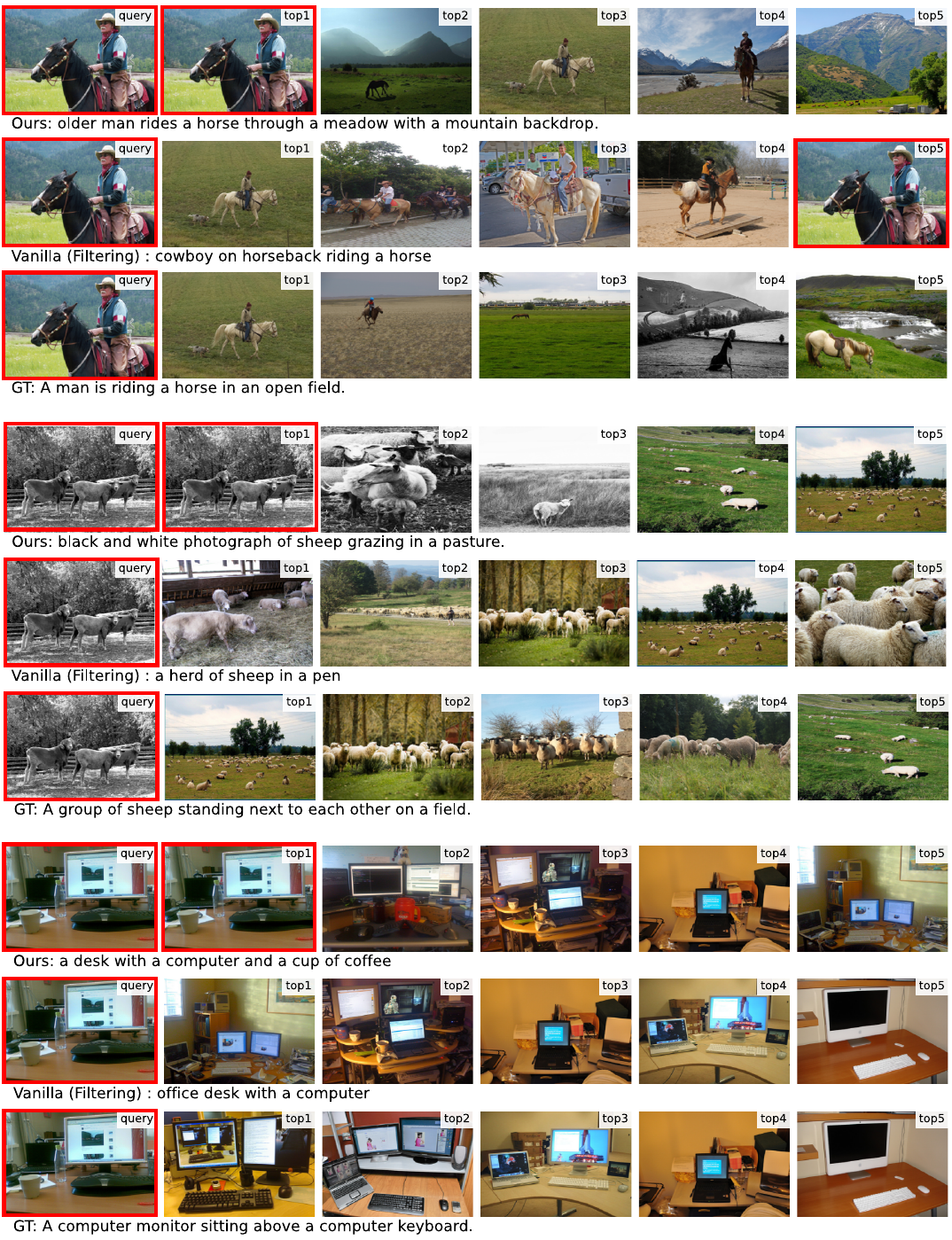}
    }
    \caption{
        \small An example of self-retrieval in MSCOCO. In the example, the first column indicates the input image and the generated captions by the specified model, while 2-6th columns show the top-5 retrieved images using the generated captions---by our method and Vanilla (Filtering) baseline---or ground-truth caption.
        Fine details captured by our model can enhance the search results.
    }
    \label{fig:self_retrieval_examples}
\end{figure*}

\subsection{Qualitative Analysis}
\label{sec:qualitative_results}

\paragraph{Captioning results.}
\cref{fig:qualitative} presents generated captions from Vanilla, Vanilla (Filtering), and our model with $z=\{3,5,7\}$.
Our model successfully generates captions of different quality by adjusting a control signal ($z$); when feeding $z$ corresponding to higher alignment levels ($3 \rightarrow 5 \rightarrow 7$), captions become more descriptive and distinct by capturing finer-level concepts (\eg, pasture, storm cloud, cluttered desk) from images.
In contrast, two baselines generate descriptive but less distinct captions as they typically rely on common words (or phrases) or capture only salient regions.
Note that more examples are presented in  \cref{sec:appendix:more_qualitative_results}.

\vspace{-0.2cm}
\paragraph{Self-retrieval results.}
We present examples of self-retrieval results on the MSCOCO dataset for Vanilla (Filtering) baseline and our method in \cref{fig:self_retrieval_examples}. For these models, we generate a caption from an input image (1st column) and retrieve the top-5 images (2-6th columns) using the generated caption. In addition, we also present the retrieval result using ground-truth captions for the given images. Our model successfully captures the main concept (\ie, riding a horse) as well as fine details (e.g., older man, mountain backdrop) that is not captured by the baseline and even in the ground-truth caption. Such captured fine details allow us to search for more similar images for a given image as well as improve the self-retrieval performance. More examples are provided in \cref{sec:appendix:more_qualitative_results}.

%% file: sec/5_conclusion.tex

\section{Conclusion}
The recent introduction of large-scale web-crawled data has brought remarkable advances in various computer vision tasks, such as image captioning. However, since the web-crawled data relies on alt-texts, not human annotations, it inevitably includes noisy pairs in a high ratio.
Most of the recent works that use large-scale data, however, train models without any consideration for handling the noisy pairs except for simple rule-based filtering strategies.
In this paper, we first argue the importance of handling the noise issue in web-crawled image-text data, especially for image captioning.
From the comprehensive experiments, our proposed noise-aware learning framework consistently outperforms other carefully designed baselines. 
We hope our study provides insights to further explore an effective noise-aware learning algorithm for handling inherent noises of a large-scale web-crawled dataset in the future.

%% file: sec/6_appendix.tex
\appendix

\section*{Appendix}
\section{Training Details for All Experiments}
\label{sec:appendix:training_details}
We provide detailed explanations for experimental settings of all our reported results in the main paper. We have conducted our experiments under two different settings: 1) \textit{Learnable-Dense}, 2) \textit{Frozen-CLS}. The main difference between the two configurations depends on the way of utilizing the visual encoder. Regarding the text decoder, we utilize the same decoder model, which consists of a 6-layer transformer network initialized randomly for both settings.

For the main experiments \ie,~\cref{table:zero_cc,table:sota_comp,table:self_retrieval}, we use the \textit{Learnable-Dense} setting where the visual encoder, \ie, pre-trained CLIP ViT-L/14, is tunned during the training phase and both the output [CLS] feature and other spatial features of the CLIP visual encoder are used as visual features. Since the length of output spatial visual features is very long ($16 \times 16=256$), we apply 2d-average pooling to the spatial features with a scale factor of 1/4. As a result, the total length of the visual features becomes 17 ($1 + 4 \times 4$). As optimization hyperparameters, we set learning rates to 1$e$-5 and 1$e$-4 for the visual encoder and the text decoder, respectively, with the same weight decay of 1$e$-5. 

On the other hand, we adopt the \textit{Frozen-CLS} setting for all other analysis and ablation experiments due to its training efficiency. In the \textit{Frozen-CLS} setting, the visual encoder, \ie. pre-trained CLIP ViT-L/14, is frozen and only the output [CLS] feature is used as the visual feature. In this setting, since the sequence length of the visual feature is 1, we use the visual feature as a prefix token of caption tokens like self-attention-based decoding in GIT~\cite{git}. We set the learning rate for the caption decoder to 0.0016. 

For both settings, we train our model for 10 epochs. The learning rate is warmed up in the first epoch and then follows cosine decay to 0. Also, the network parameters are updated by AdamW~\cite{adamw} with $\beta_1$ = 0.9 and $\beta_2$ = 0.999.


\section{Network Architecture Details}
\label{sec:appendix:network_arch}
\begin{figure}[t]
    \begin{center}
    \scalebox{0.95}{
        \includegraphics[width=\linewidth]{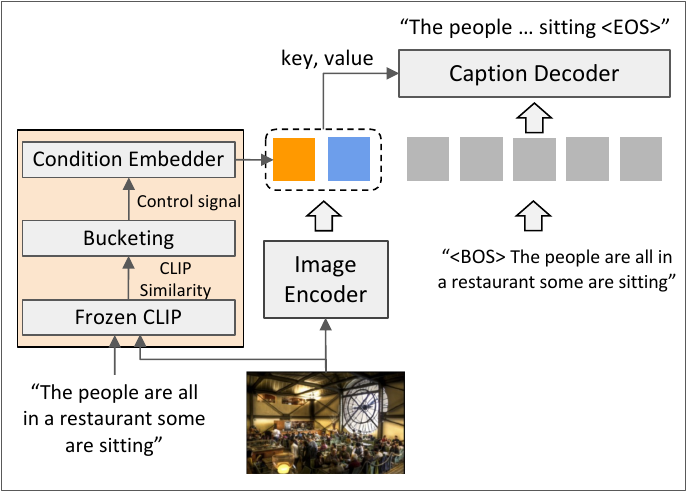}
    }
    \vspace{-0.3cm}
    \end{center}
    \caption{
        \small The network architecture of our alignment-level-controllable captioner. During the training phase, the calculated control signal is concatenated to the image embeddings. Then the concatenated vectors are fed into a cross-attention-based caption decoder model as key-value features.
    }
    \label{fig:net_arch}
    \vspace{-0.35cm}
\end{figure}

We present the network architecture of our alignment-level-controllable captioner in~\cref{fig:net_arch}. 
When a pair of an image and a caption is given, we first calculate the cosine similarity, $s \in \mathbb{R}$, of the pair using pre-trained CLIP ViT-L/14. Then, we convert the similarity score into a discrete alignment level $l \in \{ 1, ..., K \}$ via the bucketing technique as described in~\cref{sec:method:noise_acquisition}. 
After getting the alignment level, we feed the discrete alignment level as a control signal into a learnable embedding layer to get a control vector that has the same dimension as the image embedding extracted from an image encoder. 
Finally, we concatenate the control signal and image embedding vectors and feed them to a caption decoder model. During the inference phase, we can simply set the control signal $z$ as the desired alignment level (\eg, $z=7$) to get a caption describing a given image.

\begin{table}[t]
	\centering
        \caption{
		\small Zero-shot results on MSCOCO with different decoder architectures trained on CC3M dataset.
	}
	\vspace{-0.35cm}
	\scalebox{0.75}{
		\begin{tabular}{ll|ccccc}
			\toprule
			Decoder & Method  & B$@$4 & METEOR & SPICE & CIDEr \\
			\hline\hline
			\multirow{3}{*}{GiT-like~\cite{git}} & Vanilla    & 8.65 & 14.40 & 10.32 & 40.24 \\
			& Filtering  & 11.00 & 16.44 & 11.64 & 47.75 \\
			& NoC (z=7) & \textbf{12.70} & \textbf{18.05} & \textbf{12.95} & \textbf{51.11} \\
			\hline
			\multirow{3}{*}{VirTex-like~\cite{virtex}} & Vanilla     & 9.37 & 15.02 & 10.75 & 42.43\\
			& Filtering   & 11.24 & 16.89 & 12.37 & 50.43\\
			& NoC (z=7)  & \textbf{13.42} & \textbf{18.73} & \textbf{13.49} & \textbf{53.18} \\
			\bottomrule
		\end{tabular}
	}
	\vspace{-0.4cm}
	\label{tab:ablation_decoder}
\end{table}

\section{Backbone Agnostic Property}
\label{sec:appendix:backbone_agnostic_property}
Our method requires only a minor modification (\ie, adding a control signal) to a conventional image captioning model, which means that our proposed noise-aware learning framework can be easily applied to any captioning model.
Despite we have mainly reported the results based on a VirTex-like~\cite{virtex} cross-attention-based transformer network in our main paper, our noise-aware learning framework also can be simply applied to a GIT-like~\cite{git} self-attention-based transformer network. For the VirTex-like architecture, the control signal is concatenated to visual features and fed into each cross-attention layer as key-value features like~\cref{fig:net_arch}. While in the GIT-like architecture, we can simply use the concatenated features as prefix tokens of a caption. 

To empirically validate the backbone agnostic property of our algorithm, we present additional comparative experiments in~\cref{tab:ablation_decoder}. In this experiment, we freeze the visual encoder and use both [CLS] token and other spatial features, referred to as \textit{Frozen-Dense}, for training efficiency. From the~\cref{tab:ablation_decoder}, our proposed model consistently outperforms fairly-controlled comparative baselines by large margins on both decoder architectures.


\begin{table*}[!t]
    \begin{center}
    \scalebox{0.90}{
    \begin{tabular}{l|cccc|cc|cc|cc|cc}
        \toprule 
        \multirow{3}{*}{Models} &\multicolumn{4}{c|}{\multirow{2}{*}{MSCOCO}} &\multicolumn{8}{c}{nocaps} \\
        & & & & &\multicolumn{2}{c|}{overall} &\multicolumn{2}{c|}{in-domain} &\multicolumn{2}{c|}{near-domain} &\multicolumn{2}{c}{out-of-domain} \\
        &B@4 &M &C &CS &C &CS &C &CS &C &CS &C &CS \\

        \midrule\midrule
        Vanilla &10.31 &15.48 &47.56 &62.89 &41.58 &60.49 &38.60 &58.64 &39.24 &59.91 &51.22 &62.15 \\
        Vanilla (Filtering) &12.81 & 17.30 & 54.66 & 64.84 & 48.96 & 62.70 & 46.06 & 60.74 & 46.33 & 62.35 & 59.50 & 63.92 \\
        Bootstrap & 13.51 & 17.46 & 55.13 & 64.31 & 49.46 & 62.16 & 45.23 & 60.60 & 47.14 & 61.62 & 59.93 & 63.62 \\

        \midrule
        NoC (z=7) & \textbf{15.96} & \textbf{19.50} & \textbf{62.04} & \textbf{66.70} & \textbf{54.94} & \textbf{64.21} & \textbf{51.74} & \textbf{62.54} & \textbf{53.09}  & \textbf{63.92} & \textbf{63.15} & \textbf{65.19} \\
        \bottomrule
    \end{tabular}
    }
    \end{center}
    \vspace{-0.55cm}
    \caption{
        \small Comparison with the Bootstrap baseline. All models are trained on CC3M and evaluated on MSCOCO and nocaps datasets.
        B@4, M, C, and CS mean BLEU@4, METEOR, CIDEr, and CLIPScore metrics, respectively.
        Numbers in \textbf{bold} indicate the best method.
    }
    \vspace{-0.3cm}
    \label{table:zero_cc_with_bootstrap}
\end{table*}

\section{Details for Loss Weighting Baseline}
\label{sec:appendix:loss_weighting}
We have defined the Loss weighting baseline in~\cref{sec:expr_baselines} as follows:
\begin{equation}
\label{eq:loss_weight_appendix}
    \begin{split}
        \mathcal{L_{\text{weighting}}} = -{1 \over N}\sum_{i=1}^{N}{s_{i}\log p(c_{i}|I_{i})},
    \end{split}
\end{equation}
where $s_i$ indicates a cosine similarity computed by CLIP for $i^{\text{th}}$ image-text pair in a minibatch of $N$ instances.

\begin{figure}[!t]
    \begin{center}
    \begin{minipage}[b]{0.95\linewidth}
        \begin{center}
        \scalebox{0.95}{
            \includegraphics[width=\linewidth]{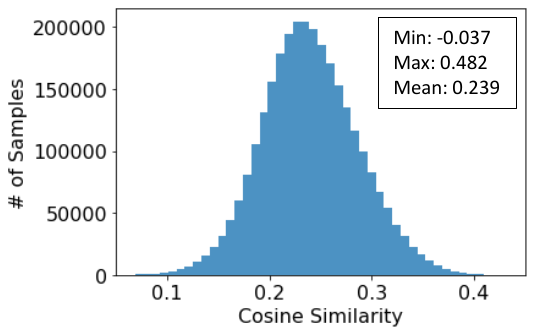}
        }
        \vspace{-0.4cm}
        \caption*{
            \small (a) Distribution of cosine similarities \textbf{\textit{before}} re-scaling.
        }
        \vspace{0.3cm}
        \end{center}
        \label{fig:cos_nonorm}
    \end{minipage}
    \begin{minipage}[b]{0.95\linewidth}
        \begin{center}
        \scalebox{0.95}{
            \includegraphics[width=\linewidth]{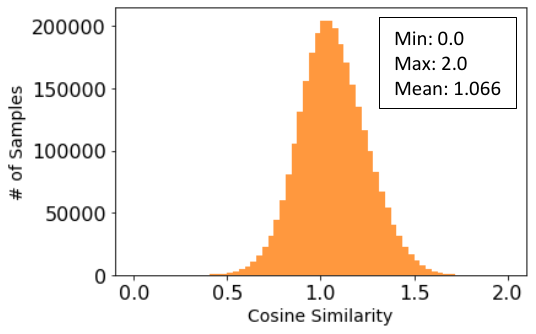}
        }
        \vspace{-0.4cm}
        \caption*{
            \small (b) Distribution of cosine similarities \textbf{\textit{after}} re-scaling.
        }
        \end{center}
        \label{fig:cos_norm}
    \end{minipage}
    \end{center}
    \vspace{-0.4cm}
    \caption{
        \small Distributions of cosine similarities for all image-text pairs in the CC3M training split.
        The cosine similarity is calculated by pre-trained CLIP ViT-L/14. Statistics for each distribution are presented in the upper right region in each figure. 
    }
    \label{fig:cos_distributions}
\end{figure}

Additionally, since the range of cosine similarity from training samples is approximately ranged between 0 and 0.5 as shown in~\cref{fig:cos_distributions}, we apply min-max normalization to s and multiply by 2 for equalizing the averaged loss scale. This loss reweighting strategy makes the model be updated by less for miss-aligned samples and more on well-aligned ones.

For a better understanding, we visualize the distribution of cosine similarities of the CC3M training split in~\cref{fig:cos_distributions}. The range of raw cosine similarities is distributed from -0.037 to 0.482, and the mean of the distribution is 0.239. 
If we use the cosine similarities directly for loss re-weighting defined in~\cref{eq:loss_weight_appendix}, the scale of the overall loss value becomes low, which could result in slow convergence of the training phase.
Thus, to equalize the averaged loss scale, we re-scale the cosine similarities for the loss re-weighting strategy by applying min-max normalization and multiplying by 2. Statistics of the resulting distribution of the re-scaled cosine similarities are 0.0, 2.0, and 1.066 for minimum, maximum, and mean values, respectively.

\begin{table*}[!t]
    \begin{center}
    \scalebox{0.90}{
    \begin{tabular}{l|cccc|cc|cc|cc|cc}
        \toprule 
        \multirow{3}{*}{Models} &\multicolumn{4}{c|}{\multirow{2}{*}{MSCOCO}} &\multicolumn{8}{c}{nocaps} \\
        & & & & &\multicolumn{2}{c|}{overall} &\multicolumn{2}{c|}{in-domain} &\multicolumn{2}{c|}{near-domain} &\multicolumn{2}{c}{out-of-domain} \\
        &B@4 &M &C &CS &C &CS &C &CS &C &CS &C &CS \\

        \midrule\midrule
        Vanilla &10.31 &15.48 &47.56 &62.89 &41.58 &60.49 &38.60 &58.64 &39.24 &59.91 &51.22 &62.15 \\
        Vanilla (Filtering) &12.81 & 17.30 & 54.66 & 64.84 & 48.96 & 62.70 & 46.06 & 60.74 & 46.33 & 62.35 & 59.50 & 63.92 \\
        Loss weighting & 11.16 & 16.15 & 50.86 & 63.87 & 43.89 & 61.18 & 39.30 & 59.23 & 41.80 & 60.50 & 53.84 & 63.04 \\
        \midrule
        NoC (z=1) & 12.65 & 16.44 & 52.78 & 63.23 & 43.95 & 60.55 & 41.39 & 58.50 & 42.11 & 60.30 & 51.70 & 61.62 \\
        NoC (z=2) & 3.77 & 8.21 & 12.83 & 45.24 & 10.26 & 43.43 & 11.96 & 46.29 & 10.76 & 44.46 & 7.45 & 40.60 \\
        NoC (z=3) & 3.70 & 9.19 & 17.33 & 49.67 & 15.12 & 48.05 & 16.45 & 49.56 & 15.68 & 48.66 & 12.40 & 46.48 \\
        NoC (z=4) & 8.19 & 13.52 & 39.61 & 59.86 & 32.25 & 57.23 & 30.04 & 56.01 & 31.76 & 57.23 & 35.39 & 57.62 \\
        NoC (z=5) & 11.88 & 16.28 & 52.11 & 64.01 & 45.68 & 61.72 & 41.25 & 59.57 & 43.74 & 61.18 & 55.05 & 63.23 \\
        NoC (z=6) & \underline{14.38} & \underline{18.27} & \underline{58.76} & \underline{65.82} & \underline{51.50} & \underline{63.52} & \underline{48.02} & \underline{61.82} & \underline{49.60} & \underline{63.13} & \underline{60.06} & 63.52 \\
        NoC (z=7) & \textbf{15.96} & \textbf{19.50} & \textbf{62.04} & \textbf{66.70} & \textbf{54.94} & \textbf{64.21} & \textbf{51.74} & \textbf{62.54} & \textbf{53.09}  & \textbf{63.92} & \textbf{63.15} & \textbf{65.19} \\
        NoC (z=8) & 12.82 & 17.46 & 53.50 & 64.94 & 48.05 & 62.64 & 42.44 & 60.16 & 45.86 & 62.20 & 59.08 & \underline{64.21} \\
        \bottomrule
    \end{tabular}
    }
    \end{center}
    \vspace{-0.55cm}
    \caption{
        \small Zero-shot captioning results for all bin indices from models trained on CC3M. Numbers in \textbf{bold} and \underline{underlined} indicate the best and second-best ones, respectively.
    }
    \vspace{-0.3cm}
    \label{table:zero_cc_all_bins}
\end{table*}

\section{Comparison with Data Bootstrapping}
\label{sec:appendix:data_bootstrapping}
While BLIP~\cite{blip} and our method have a similar motivation, \ie, handle the noise issue inherent in the web-crawled dataset, we remark that BLIP requires clean data (\eg, MSCOCO) to learn a filter and captioner for bootstrapping as we discussed in~\cref{sec:related_work:noisy_labels}. 
Despite the necessity of clean data in BLIP, we carefully devised an additional experiment to evaluate the effectiveness of the data bootstrapping technique introduced in BLIP for our zero-shot experimental setting.

Firstly, as the captioner and filter models of BLIP are fine-tuned using the clean MSCOCO dataset, we implement a replacement for the purpose of zero-shot evaluation on MSCOCO. Specifically, we employ our Vanilla (Filtering) model, which has been trained on the filtered CC3M dataset, to replace the captioner, and we utilize the pre-trained CLIP ViT-B/32 as a substitute for the filter.
Subsequently, we partitioned the original CC3M dataset into two groups. The first group consists of image-text pairs with a CLIP similarity greater than 0.3, which are considered well-aligned annotations. While the second group comprises image-text pairs with a CLIP similarity lower than 0.3, which are considered less-aligned annotations. 
After separating the dataset, we use the trained Vanilla (Filtering) model to generate pseudo-captions for the images in the second group, as suggested by BLIP.
Finally, we train a Vanilla model with the bootstrapped dataset under the \textit{Learnable-Dense} setting. 

From the~\cref{table:zero_cc_with_bootstrap}, we observed that the Bootstrap baseline shows marginally higher captioning scores than the Vanilla (Filtering). In contrast, the Bootstrap baseline shows significantly lower scores than our proposed model. We carefully hypothesize that the bootstrapping technique would yield optimal results when a clean dataset of the target domain is available for training the captioner and filter. On the other hand, our proposed model can effectively mitigate the negative impact of noisy pairs utilizing only the web-crawled data. Consequently, it shows superior zero-shot generalization performance in comparison to the data bootstrapping technique.


\section{Detailed Results for All Alignment Levels}
\label{sec:appendix:all_alignment_levels}
We provide comprehensive results for zero-shot captioning and self-retrieval tasks originally reported in~\cref{table:zero_cc,table:self_retrieval}, with the aim of using these expended results to help verify that our noise-aware model is trained according to our intentions. 
\cref{table:zero_cc_all_bins,table:zero_coyo} present zero-shot captioning results. Each model is trained on either CC3M or various scales of COYO and evaluated on MSCOCO and nocaps datasets. 
From the tables, our model consistently outperforms other baselines when ($z \ge 6$). On the other hand, when ($z < 6$), our model shows lower scores than the comparative baselines. This trend is reasonable considering that image-text pairs with bin indices of $z < 6$ are relatively less-aligned pairs. Consequently, our model, when conditioned on these indices, is expected to generate less-aligned captions.

From the~\cref{table:self_retrieval_all_bins,table:self_retrieval_coyo}, we observe a similar trend in retrieval performance as in captioning results, \ie, increasing the bin index leads to higher scores. Notably, as illustrated in~\cref{table:self_retrieval_coyo}, self-retrieval performance consistently improves as models are trained on larger-scale datasets. This suggests that a model can generate more distinctive captions as it learns a broader range of visual concepts from larger datasets.

We note that a model trained in CC3M shows irregular performance at the lowest bin index ($z=1$). We hypothesize that this is due to the extremely small number of samples for $z=1$ in CC3M (only 21 samples included), which prevents the model from being fully trained for the noisiest samples. In contrast, for COYO, which is much noisier than CC3M, the number of noisiest samples is sufficiently large, resulting in the model showing the worst performance at $z=1$.

\begin{table}[!t]
    \begin{center}
    \scalebox{0.80}{
    \begin{tabular}{l|rrr|rrr}
        \toprule
        \multirow{2}{*}{Models} &\multicolumn{3}{c|}{MSCOCO} &\multicolumn{3}{c}{Flickr30k} \\
        &R@1 &R@5 &R@10 &R@1 &R@5 &R@10 \\

        \midrule
        \textcolor{light-gray}{GT Caption} &\textcolor{light-gray}{34.57} &\textcolor{light-gray}{59.30} &\textcolor{light-gray}{69.91} &\textcolor{light-gray}{63.08} &\textcolor{light-gray}{86.50} &\textcolor{light-gray}{92.00} \\
        
        \midrule
        Vanilla & 25.44 & 50.38 & 61.66 & 47.10 & 76.60 & 85.90 \\
        Vanilla (Filtering) & 31.64 & 58.90 & 70.36 & 56.50 & 85.50 & 92.50 \\
        Loss weighting & 28.78 & 54.44 & 65.44 & 48.00 & 78.90 & 87.50 \\

        \midrule
        NoC (z=1) & 25.92 & 49.92 & 61.98 & 44.60 & 77.20 & 86.60 \\
        NoC (z=2) & 2.44 & 7.86 & 11.46 & 8.30 & 21.50 & 29.20 \\
        NoC (z=3) & 5.02 & 13.12 & 18.98 & 12.70 & 30.90 & 40.00 \\
        NoC (z=4) & 17.34 & 38.86 & 50.30 & 32.30 & 66.70 & 75.30 \\
        NoC (z=5) & 29.02 & 54.32 & 66.26 & 50.30 & 84.10 & 91.10 \\
        NoC (z=6) & \underline{35.26} & \underline{62.78} & \underline{73.88} & \underline{60.00} & \underline{89.90} & \underline{95.40} \\
        NoC (z=7) & \textbf{40.00} & \textbf{66.78} & \textbf{77.53} & \textbf{65.10} & \textbf{92.00} & \textbf{96.20} \\
        NoC (z=8) & 32.30 & 57.60 & 69.72 & 54.90 & 85.80 & 93.50 \\
        \bottomrule
    \end{tabular}
    }
    \end{center}
    \vspace{-0.55cm}
    \caption{
        \small Self-retrieval capability for all bin indices on MSCOCO and Flickr30k datasets.
        Numbers in \textbf{bold} and \underline{underlined} indicate the best and second-best ones, respectively.
    }
    \vspace{-0.35cm}
    \label{table:self_retrieval_all_bins}
\end{table}

\begin{figure*}[!t]
    \begin{center}
    \scalebox{0.95}{
        \includegraphics[width=\linewidth]{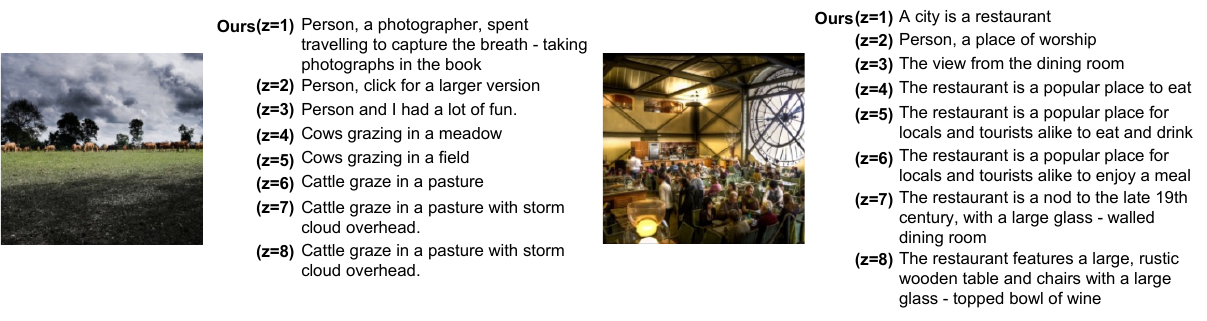}
    }
    \vspace{-0.2cm}
    \end{center}
    \caption{
        \small Examples of generated captions from all alignment levels of our model.
    }
    \label{fig:all_bin_caps}
\end{figure*}

\section{Discussion on Similarity Computation Module}
\label{sec:appendix:similarity_computation_module}
We examine how robust our model is to the variations on the alignment level computation module used for computing cosine similarities between given image-text pairs.
We change the alignment level computation module from pre-trained CLIP ViT-L/14 to CLIP ViT-B/32 which has fewer parameters resulting in faster computations of alignment levels but shows lower performance than the ViT-L/14-based model in other downstream tasks~\cite{CLIP}. The zero-shot captioning and self-retrieval results are presented in~\cref{table:zero_caption_vitb} and ~\cref{table:zero_retrieval_vitb}, respectively.
From the tables, we observe that our model shows almost consistent performance on both the captioning and retrieval tasks, even if the alignment level computation module is changed to a model having relatively lower representation power.
Also, we expect that if we can use a more powerful model trained to capture fine-grained alignments between images and paired texts, such as Florence~\cite{florence} or FILIP~\cite{FILIP}\footnote{Checkpoints of the models have not been released publicly yet.}, for measuring the alignment levels, our proposed noise-aware learning framework can perform better than current results as the powerful alignment computation module measures the alignment level more accurately.


\section{More Qualitative Results}
\label{sec:appendix:more_qualitative_results}
\subsection{Zero-shot Captioning Results}
We present the captioning results for all alignment levels of our model to show that our model can effectively control the quality of generated captions in~\cref{fig:all_bin_caps}. 
From the~\cref{fig:all_bin_caps}, our quality controllable model generates miss aligned or less descriptive captions with the control signals corresponding to low alignment levels (\ie, $z \le 3$). 
While, with the control signals meaning to middle alignment levels (\ie, $4 \le z \le 6$), our model describes the images using common words as other baselines do.
Finally, our model with the control signals corresponding to high alignment levels (\ie, $z \ge 7$) can generate captions satisfying both \textit{descriptiveness} and \textit{distinctiveness}. 

\begin{figure*}[!t]
    \centering
    \scalebox{0.95}{
        \includegraphics[width=\linewidth]{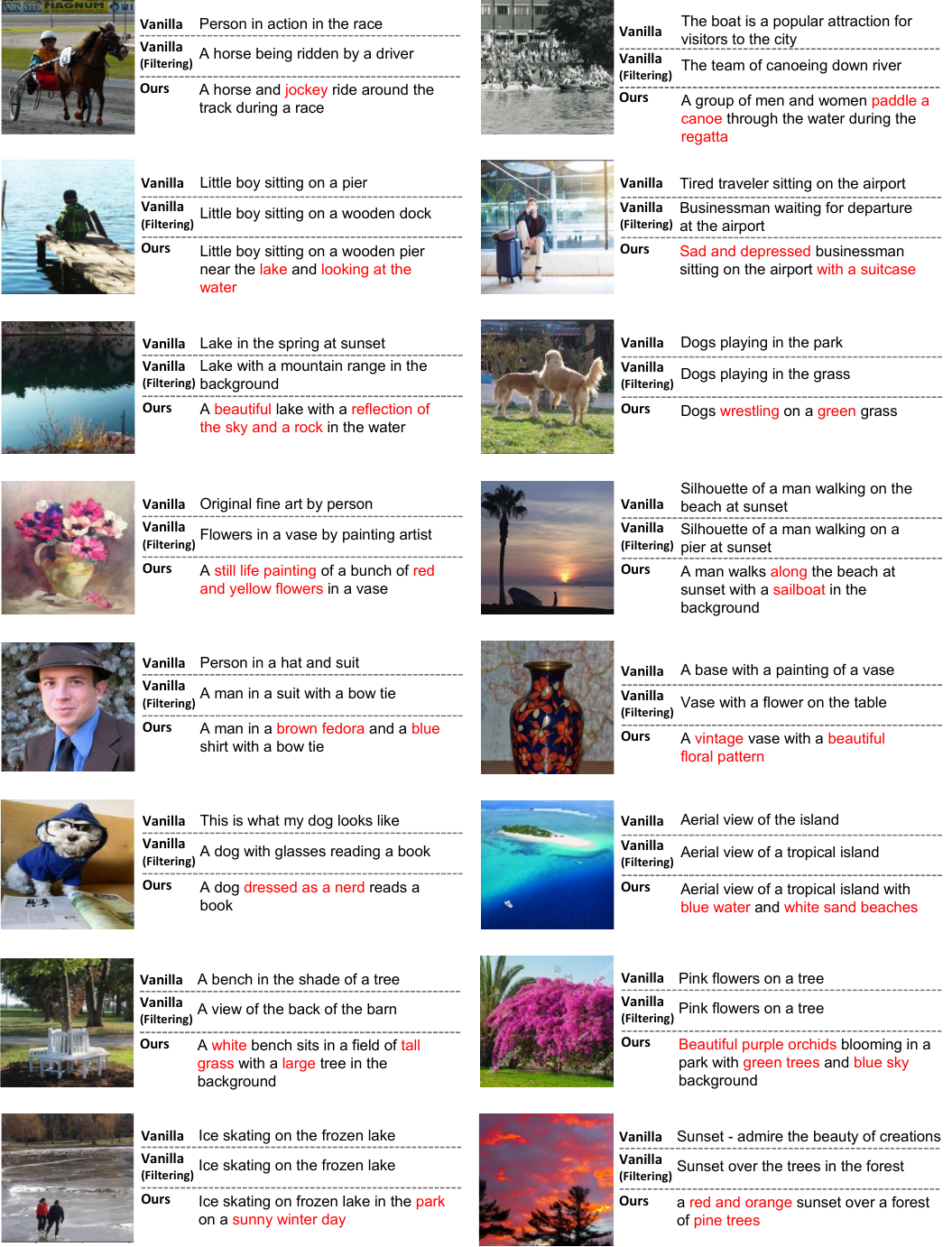}
    }
    \vspace{-0.2cm}
    \caption{
        \small Examples of generated captions sampled from MSCOCO and CC3M validation splits. The captions of our model are generated with the control signal $z=7$.
        Expressions capturing fine details from images in ours are highlighted in red.
    }
    \label{fig:more_qualitative_results}
\end{figure*}

In addition, we provide more zero-shot captioning examples in~\cref{fig:more_qualitative_results}. As discussed in~\cref{sec:qualitative_results}, baseline models tend to describe given images using common words (or phrases), while our model with a higher alignment level (\eg, $z=7$) describes the images with more various words (or phrases) based on learned rich visual knowledge.

\subsection{Self-retrieval Results}
\begin{figure*}[!t]
    \centering
    \scalebox{0.90}{
        \includegraphics[width=\linewidth]{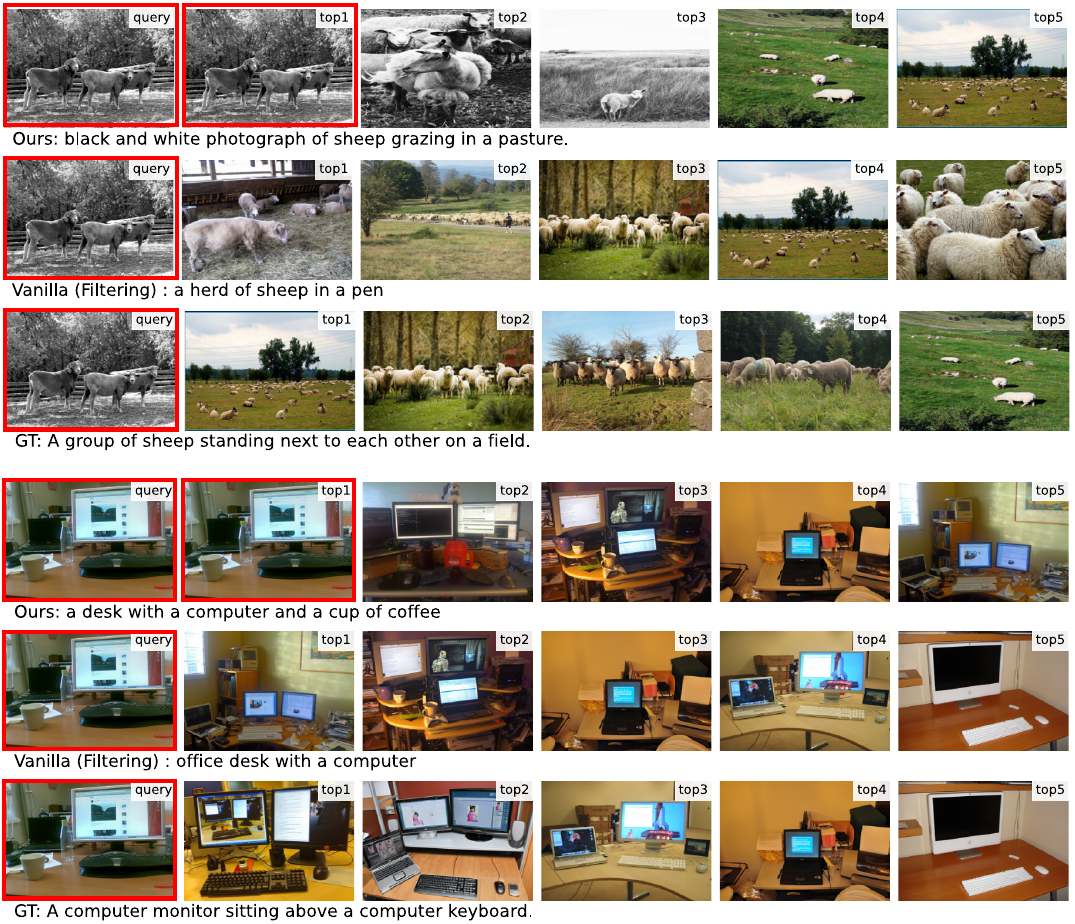}
    }
    \caption{
        \small Examples of self-retrieval in MSCOCO. For each example (of three rows), the first column indicates the input image and the generated captions by the specified model, while 2-6th columns show the top-5 retrieved images using the generated captions---by our method and Vanilla (Filtering) baseline---or ground-truth caption.
    }
    \label{fig:self_retrieval_examples_more}
\end{figure*}

We present more examples of self-retrieval results on the MSCOCO dataset for Vanilla (Filtering) baseline and our method in \cref{fig:self_retrieval_examples_more}. 


\begin{table*}[!t]
    \centering
    \begin{minipage}[b]{1.0\linewidth}
        \centering
        \scalebox{0.75}{
        \begin{tabular}{l|cccc|cc|cc|cc|cc}
            \toprule 
            \multirow{3}{*}{Models} &\multicolumn{4}{c|}{\multirow{2}{*}{MSCOCO}} &\multicolumn{8}{c}{nocaps} \\
            & & & & &\multicolumn{2}{c|}{overall} &\multicolumn{2}{c|}{in-domain} &\multicolumn{2}{c|}{near-domain} &\multicolumn{2}{c}{out-of-domain} \\
            &B@4 &M &C &CS &C &CS &C &CS &C &CS &C &CS \\
    
            \midrule\midrule
            Vanilla &4.41 &10.49 &22.54 &58.25 &21.30 &57.62 &17.76 &55.97 &18.98 &56.74 &31.26 &59.79 \\
            Vanilla (Filtering) &4.65 &11.52 &22.48 &59.67 &21.29 &59.04 &18.15 &56.22 &19.25 &58.68 &30.07 &60.56  \\
    
            \midrule
            NoC (z=1) &0.00 &2.40 &0.09 &30.86 &0.08 &30.85 &0.05 &30.72 &0.09 &30.29 &0.07 &31.95 \\
            NoC (z=2) &0.00 &2.38 &0.13 &31.20 &0.15 &31.36 &0.24 &30.17 &0.15 &30.89 &0.10 &32.60 \\
            NoC (z=3) &0.31 &4.28 &3.15 &41.82 &2.31 &40.49 &2.74 &40.33 &2.32 &40.52 &1.97 &40.47 \\
            NoC (z=4) &2.40 &8.30 &16.21 &54.88 &17.05 &54.33 &13.20 &52.25 &15.45 &53.53 &24.93 &56.44 \\
            NoC (z=5) &5.40 &11.37 &25.80 &60.35 &24.21 &59.36 &20.17 &56.96 &21.84 &58.81 &34.70 &61.10 \\
            NoC (z=6) &\textbf{6.60} &13.39 &\textbf{28.90} &62.50 &\underline{26.92} &61.22 &\textbf{23.19} &58.33 &\underline{24.84} &61.00 &\underline{36.24} &62.48 \\
            NoC (z=7) &\underline{6.59} &\underline{14.19} &\underline{28.10} &\textbf{63.33} &\textbf{27.24} &\textbf{62.09} &\underline{22.17} &\textbf{59.32} &\textbf{25.29} &\textbf{61.86} &\textbf{37.09} &\textbf{63.34} \\
            NoC (z=8) &6.06 &\textbf{14.39} &24.72 &\underline{63.13} &24.85 &\underline{61.88} &19.57 &\underline{58.67} &23.14 &\underline{61.63} &34.11 &\underline{63.31} \\
            \bottomrule
            \end{tabular}
            \label{table:zero_coyo3m}
        }
        \vspace{-0.25cm}
        \caption*{
            \small (a) Zero-shot captioning results from models trained on COYO3M
        }
        \vspace{0.1cm}
    \end{minipage}
    \begin{minipage}[b]{1.0\linewidth}
        \centering
        \scalebox{0.75}{
            \begin{tabular}{l|cccc|cc|cc|cc|cc}
                \toprule 
                \multirow{3}{*}{Models} &\multicolumn{4}{c|}{\multirow{2}{*}{MSCOCO}} &\multicolumn{8}{c}{nocaps} \\
                & & & & &\multicolumn{2}{c|}{overall} &\multicolumn{2}{c|}{in-domain} &\multicolumn{2}{c|}{near-domain} &\multicolumn{2}{c}{out-of-domain} \\
                &B@4 &M &C &CS &C &CS &C &CS &C &CS &C &CS \\
        
                \midrule\midrule
                Vanilla &5.34 &11.31 &27.52 &60.69 &24.00 &59.61 &18.32 &56.52 &21.95 &58.96 &34.59 &61.77 \\
                Vanilla (Filtering) &6.51 &12.71 &29.25 &64.11 &26.88 &62.83 &21.39 &60.87 &24.84 &62.52 &37.30 &63.98 \\
        
                \midrule
                NoC (z=1) &0.10 &2.37 &0.13 &28.44 &0.12 &27.39 &0.17 &26.64 &0.11 &27.17 &0.11 &28.03 \\
                NoC (z=2) &0.00 &2.44 &0.28 &32.28 &0.23 &31.75 &0.29 &30.62 &0.26 &31.63 &0.12 &32.30 \\
                NoC (z=3) &0.52 &4.33 &3.56 &41.89 &2.46 &40.45 &2.85 &40.32 &2.49 &40.39 &2.08 &40.60 \\
                NoC (z=4) &2.90 &8.94 &19.62 &56.20 &18.78 &55.61 &15.70 &52.46 &16.80 &54.92 &27.33 &57.84 \\
                NoC (z=5) &6.22 &12.09 &29.86 &62.60 &27.27 &61.44 &21.82 &58.53 &25.45 &61.04 &36.97 &63.08 \\
                NoC (z=6) &\textbf{7.32} &14.05 &\textbf{32.63} &65.33 &\underline{30.57} &63.91 &\underline{25.33} &60.63 &\textbf{28.28} &63.77 &41.65 &65.12 \\
                NoC (z=7) &\underline{7.05} &\underline{15.03} &\underline{30.59} &\underline{66.26} &\textbf{30.66} &\underline{65.08} &\textbf{26.11} &\underline{61.95} &\underline{27.91} &\underline{64.94} &\textbf{42.72} &\underline{66.27} \\
                NoC (z=8) &6.92 &\textbf{15.59} &27.75 &\textbf{66.65} &29.07 &\textbf{65.35} &24.75 &\textbf{63.06} &26.02 &\textbf{65.20} &\underline{41.93} &\textbf{66.32} \\
                \bottomrule
            \end{tabular}
            \label{table:zero_coyo10m}
        }
        \vspace{-0.25cm}
        \caption*{
            \small (b) Zero-shot captioning results from models trained on COYO10M
        }
        \vspace{0.1cm}
    \end{minipage}
    \begin{minipage}[b]{1.0\linewidth}
        \centering
        \scalebox{0.75}{
            \begin{tabular}{l|cccc|cc|cc|cc|cc}
                \toprule 
                \multirow{3}{*}{Models} &\multicolumn{4}{c|}{\multirow{2}{*}{MSCOCO}} &\multicolumn{8}{c}{nocaps} \\
                & & & & &\multicolumn{2}{c|}{overall} &\multicolumn{2}{c|}{in-domain} &\multicolumn{2}{c|}{near-domain} &\multicolumn{2}{c}{out-of-domain} \\
                &B@4 &M &C &CS &C &CS &C &CS &C &CS &C &CS \\
        
                \midrule\midrule
                Vanilla &5.17 &11.32 &27.07 &62.06 &25.32 &60.93 &19.29 &58.43 &23.05 &60.26 &36.87 &62.95 \\
                Vanilla (Filtering) &7.18 &13.33 &\underline{32.52} &65.87 &29.85 &64.65 &\underline{25.33} &62.48 &27.64 &64.25 &40.18 &66.05 \\
        
                \midrule
                NoC (z=1) &0.00 &2.96 &0.08 &33.45 &0.08 &31.54 &0.11 &30.90 &0.08 &31.34 &0.07 &32.12 \\
                NoC (z=2) &0.00 &3.14 &0.18 &35.89 &0.18 &34.65 &0.26 &33.81 &0.19 &34.39 &0.08 &35.38 \\
                NoC (z=3) &0.36 &4.33 &2.57 &41.26 &1.68 &39.40 &1.88 &39.34 &1.78 &39.70 &1.20 &38.86 \\
                NoC (z=4) &2.81 &9.03 &19.97 &56.93 &18.84 &56.11 &14.33 &53.29 &17.19 &55.30 &27.34 &58.46 \\
                NoC (z=5) &6.11 &12.05 &30.25 &63.67 &27.29 &62.53 &22.40 &59.81 &25.11 &62.06 &37.75 &64.21 \\
                NoC (z=6) &\textbf{7.59} &14.19 &\textbf{33.43} &66.65 &\textbf{30.60} &65.17 &\textbf{26.59} &62.64 &\underline{28.11} &65.00 &\textbf{41.45} &66.24 \\
                NoC (z=7) &\underline{7.57} &\underline{15.54} &31.72 &\underline{68.12} &\underline{30.55} &\underline{66.59} &25.25 &\underline{64.00} &\textbf{28.52} &\underline{66.46} &\underline{40.85} &\underline{67.60} \\
                NoC (z=8) &7.01 &\textbf{16.37} &25.45 &\textbf{68.80} &27.11 &\textbf{67.39} &21.02 &\textbf{64.92} &24.34 &\textbf{67.26} &40.33 &\textbf{68.36} \\
                \bottomrule
            \end{tabular}
            \label{table:zero_coyo23m}
        }
        \vspace{-0.25cm}
        \caption*{
            \small (c) Zero-shot captioning results from models trained on COYO23M
        }
        \vspace{0.1cm}
    \end{minipage}
    \begin{minipage}[b]{1.0\linewidth}
        \centering
        \scalebox{0.75}{
            \begin{tabular}{l|cccc|cc|cc|cc|cc}
                \toprule 
                \multirow{3}{*}{Models} &\multicolumn{4}{c|}{\multirow{2}{*}{MSCOCO}} &\multicolumn{8}{c}{nocaps} \\
                & & & & &\multicolumn{2}{c|}{overall} &\multicolumn{2}{c|}{in-domain} &\multicolumn{2}{c|}{near-domain} &\multicolumn{2}{c}{out-of-domain} \\
                &B@4 &M &C &CS &C &CS &C &CS &C &CS &C &CS \\
        
                \midrule\midrule
                Vanilla &4.92 &11.18 &28.03 &62.40 &25.29 &61.73 &20.25 &59.14 &22.56 &60.98 &37.63 &63.90 \\
                Vanilla (Filtering) &7.80 &13.48 &\underline{34.55} &66.75 &30.93 &65.56 &\underline{26.90} &63.47 &28.75 &65.35 &40.77 &66.58 \\
        
                \midrule
                NoC (z=1) &0.00 &2.04 &0.09 &33.84 &0.09 &31.25 &0.16 &30.30 &0.08 &30.71 &0.10 &32.56 \\
                NoC (z=2) &0.00 &2.04 &0.20 &35.42 &0.18 &33.37 &0.38 &32.51 &0.15 &32.94 &0.17 &34.44 \\
                NoC (z=3) &0.45 &4.38 &2.98 &40.84 &1.96 &38.61 &2.00 &38.88 &2.15 &39.09 &1.30 &37.62 \\
                NoC (z=4) &3.11 &9.18 &21.26 &57.13 &19.15 &56.31 &15.32 &53.77 &17.37 &55.53 &27.59 &58.54 \\
                NoC (z=5) &6.68 &12.37 &31.54 &64.60 &27.62 &63.35 &22.70 &60.74 &25.68 &62.98 &37.34 &64.82 \\
                NoC (z=6) &\textbf{8.44} &14.53 &\textbf{35.82} &67.72 &\textbf{31.62} &66.32 &\textbf{27.12} &64.30 &\underline{28.98} &66.18 &\underline{43.33} &67.19 \\
                NoC (z=7) &\underline{8.08} &\underline{15.87} &32.86 &\underline{69.24} &\underline{31.55} &\underline{67.74} &24.05 &\underline{65.75} &\textbf{29.05} &\underline{67.66} &\textbf{44.95} &\underline{68.46} \\
                NoC (z=8) &7.12 &\textbf{16.73} &23.85 &\textbf{69.87} &25.35 &\textbf{68.51} &18.08 &\textbf{66.73} &22.44 &\textbf{68.41} &39.85 &\textbf{69.24} \\
                \bottomrule
            \end{tabular}
            \label{table:zero_coyo125m}
        }
        \vspace{-0.25cm}
        \caption*{
            \small (d) Zero-shot captioning results from models trained on COYO125M
        }
        \vspace{0.1cm}
    \end{minipage}
    \caption{
        \small Zero-shot caption generation performance on MSCOCO and nocaps when scaling up the training dataset sizes using COYO. 
        Models of each dataset are trained for the same number of steps.
        Numbers in \textbf{bold} and \underline{underlined} indicate the best and second-best ones, respectively.
    }
    \label{table:zero_coyo}
\end{table*}

\begin{table*}[!t]
    \centering
    \begin{minipage}[b]{0.48\linewidth}
        \centering
        \scalebox{0.80}{
            \begin{tabular}{l|rrr|rrr}
                \toprule
                \multirow{2}{*}{Models} &\multicolumn{3}{c|}{MSCOCO} &\multicolumn{3}{c}{Flickr30k} \\
                &R@1 &R@5 &R@10 &R@1 &R@5 &R@10 \\
        
                \midrule
                \textcolor{light-gray}{GT Caption} &\textcolor{light-gray}{34.57} &\textcolor{light-gray}{59.30} &\textcolor{light-gray}{69.91} &\textcolor{light-gray}{63.08} &\textcolor{light-gray}{86.50} &\textcolor{light-gray}{92.00} \\
                
                \midrule
                Vanilla &24.92 &48.96 &59.90 &47.90 &73.90 &82.10 \\
                Vanilla (Filtering) &31.78 &57.96 &69.08 &52.20 &82.00 &90.80 \\
        
                \midrule
                NoC (z=1) &0.08 &0.24 &0.44 &0.20 &1.10 &1.90 \\
                NoC (z=2) &0.08 &0.32 &0.78 &0.10 &0.90 &1.70 \\
                NoC (z=3) &1.32 &4.20 &6.50 &3.10 &10.10 &15.00 \\
                NoC (z=4) &14.16 &34.10 &44.30 &28.40 &54.90 &66.60 \\
                NoC (z=5) &30.16 &57.14 &67.86 &54.30 &81.10 &88.90 \\
                NoC (z=6) &39.28 &68.30 &77.86 &62.80 &88.30 &92.90 \\
                NoC (z=7) &\underline{44.96} &\underline{72.56} &\underline{81.42} &\underline{66.50} &\underline{89.40} &\underline{94.70} \\
                NoC (z=8) &\textbf{45.76} &\textbf{73.02} &\textbf{81.90} &\textbf{69.70} &\textbf{91.00} &\textbf{94.80} \\
                \bottomrule
            \end{tabular}
            \label{table:self_retrieval_coyo3m}
        }
        \vspace{-0.25cm}
        \caption*{
            \small (a) Self-retrieval results from models trained on COYO3M
        }
        \vspace{0.1cm}
    \end{minipage}
    \hfill
    \begin{minipage}[b]{0.48\linewidth}
        \centering
        \scalebox{0.80}{
            \begin{tabular}{l|rrr|rrr}
                \toprule
                \multirow{2}{*}{Models} &\multicolumn{3}{c|}{MSCOCO} &\multicolumn{3}{c}{Flickr30k} \\
                &R@1 &R@5 &R@10 &R@1 &R@5 &R@10 \\
        
                \midrule
                \textcolor{light-gray}{GT Caption} &\textcolor{light-gray}{34.57} &\textcolor{light-gray}{59.30} &\textcolor{light-gray}{69.91} &\textcolor{light-gray}{63.08} &\textcolor{light-gray}{86.50} &\textcolor{light-gray}{92.00} \\
                
                \midrule
                Vanilla &31.44 &55.48 &66.42 &56.30 &79.90 &86.20 \\
                Vanilla (Filtering) &43.56 &70.60 &80.42 &67.80 &90.30 &95.00 \\
        
                \midrule
                NoC (z=1) &0.06 &0.22 &0.44 &0.20 &1.00 &1.80 \\
                NoC (z=2) &0.18 &0.44 &0.94 &0.40 &1.40 &2.20 \\
                NoC (z=3) &1.48 &4.72 &7.08 &3.10 &9.70 &14.90 \\
                NoC (z=4) &16.24 &36.94 &48.14 &33.80 &60.40 &71.70 \\
                NoC (z=5) &35.98 &63.52 &75.08 &59.70 &86.60 &92.70 \\
                NoC (z=6) &48.24 &76.44 &84.82 &71.20 &92.60 &96.80 \\
                NoC (z=7) &\underline{54.70} &\underline{80.68} &\underline{87.94} &\underline{75.50} &\underline{94.30} &\textbf{97.40} \\
                NoC (z=8) &\textbf{56.60} &\textbf{81.92} &\textbf{88.76} &\textbf{78.30} &\textbf{95.40} &\textbf{97.40} \\
                \bottomrule
            \end{tabular}
            \label{table:self_retrieval_coyo10m}
        }
        \vspace{-0.25cm}
        \caption*{
            \small (b) Self-retrieval results from models trained on COYO10M
        }
        \vspace{0.1cm}
    \end{minipage}
    \begin{minipage}[b]{0.48\linewidth}
        \centering
        \scalebox{0.80}{
            \begin{tabular}{l|rrr|rrr}
                \toprule
                \multirow{2}{*}{Models} &\multicolumn{3}{c|}{MSCOCO} &\multicolumn{3}{c}{Flickr30k} \\
                &R@1 &R@5 &R@10 &R@1 &R@5 &R@10 \\
        
                \midrule
                \textcolor{light-gray}{GT Caption} &\textcolor{light-gray}{34.57} &\textcolor{light-gray}{59.30} &\textcolor{light-gray}{69.91} &\textcolor{light-gray}{63.08} &\textcolor{light-gray}{86.50} &\textcolor{light-gray}{92.00} \\
                
                \midrule
                Vanilla &34.70 &60.42 &71.02 &58.70 &83.20 &88.30 \\
                Vanilla (Filtering) &48.70 &75.62 &84.28 &74.10 &93.30 &96.30 \\
        
                \midrule
                NoC (z=1) &0.02 &0.18 &0.30 &0.00 &0.70 &0.90 \\
                NoC (z=2) &0.06 &0.34 &0.68 &0.40 &1.20 &1.90 \\
                NoC (z=3) &1.30 &3.82 &6.42 &2.90 &8.20 &13.90 \\
                NoC (z=4) &17.44 &38.68 &50.14 &33.50 &61.00 &72.00 \\
                NoC (z=5) &39.04 &66.58 &77.42 &65.90 &89.50 &94.60 \\
                NoC (z=6) &52.80 &79.42 &87.62 &76.60 &93.50 &96.90 \\
                NoC (z=7) &\underline{61.40} &\underline{85.32} &\underline{91.38} &\underline{81.30} &\underline{96.60} &\underline{98.00} \\
                NoC (z=8) &\textbf{64.92} &\textbf{86.46} &\textbf{92.60} &\textbf{84.10} &\textbf{96.80} &\textbf{98.40} \\
                \bottomrule
            \end{tabular}
            \label{table:self_retrieval_coyo23m}
        }
        \vspace{-0.25cm}
        \caption*{
            \small (c) Self-retrieval results from models trained on COYO23M
        }
        \vspace{0.1cm}
    \end{minipage}
    \hfill
    \begin{minipage}[b]{0.48\linewidth}
        \centering
        \scalebox{0.80}{
            \begin{tabular}{l|rrr|rrr}
                \toprule
                \multirow{2}{*}{Models} &\multicolumn{3}{c|}{MSCOCO} &\multicolumn{3}{c}{Flickr30k} \\
                &R@1 &R@5 &R@10 &R@1 &R@5 &R@10 \\
        
                \midrule
                \textcolor{light-gray}{GT Caption} &\textcolor{light-gray}{34.57} &\textcolor{light-gray}{59.30} &\textcolor{light-gray}{69.91} &\textcolor{light-gray}{63.08} &\textcolor{light-gray}{86.50} &\textcolor{light-gray}{92.00} \\
                
                \midrule
                Vanilla &36.20 &60.84 &71.44 &60.90 &83.20 &89.20 \\
                Vanilla (Filtering) &52.10 &77.34 &86.10 &77.20 &95.20 &97.70 \\
        
                \midrule
                NoC (z=1) &0.02 &0.12 &0.28 &0.00 &0.40 &0.80 \\
                NoC (z=2) &0.04 &0.34 &0.58 &0.10 &0.90 &1.50 \\
                NoC (z=3) &0.94 &3.56 &5.66 &2.70 &9.10 &13.50 \\
                NoC (z=4) &17.02 &37.60 &49.36 &34.40 &64.10 &75.10 \\
                NoC (z=5) &42.00 &69.54 &79.56 &68.70 &91.40 &95.70 \\
                NoC (z=6) &57.24 &82.06 &89.24 &82.60 &96.30 &98.20 \\
                NoC (z=7) &\underline{65.32} &\underline{87.68} &\underline{92.84} &\underline{85.40} &\underline{97.40} &\underline{98.50} \\
                NoC (z=8) &\textbf{69.66} &\textbf{89.28} &\textbf{94.12} &\textbf{88.20} &\textbf{98.00} &\textbf{99.30} \\
                \bottomrule
            \end{tabular}
            \label{table:self_retrieval_coyo125m}
        }
        \vspace{-0.25cm}
        \caption*{
            \small (d) Self-retrieval results from models trained on COYO125M
        }
        \vspace{0.1cm}
    \end{minipage}
    \vspace{-0.2cm}
    \caption{
        \small Self-retrieval performance on MSCOCO and Flickr30k when scaling up the training dataset sizes using COYO. 
        Models of each dataset are trained for the same number of steps.
        Numbers in \textbf{bold} and \underline{underlined} indicate the best and second-best ones, respectively.
    }
    \label{table:self_retrieval_coyo}
\end{table*}

\clearpage

\begin{table*}[!t]
    \centering
    \scalebox{0.92}{
    \begin{tabular}{l|cccc|cc|cc|cc|cc}
        \toprule 
        \multirow{3}{*}{Models} &\multicolumn{4}{c|}{\multirow{2}{*}{MSCOCO}} &\multicolumn{8}{c}{nocaps} \\
        & & & & &\multicolumn{2}{c|}{overall} &\multicolumn{2}{c|}{in-domain} &\multicolumn{2}{c|}{near-domain} &\multicolumn{2}{c}{out-of-domain} \\
        &B@4 &M &C &CS &C &CS &C &CS &C &CS &C &CS \\

        \midrule\hline
        \rowcolor[gray]{0.85}\multicolumn{13}{l}{\textit{\textbf{Alignment Level Computation using ViT-L/14}}} \\ \hline
        NoC (z=1) &3.59 &10.36 &18.33 &50.68 &13.46 &48.28 &13.69 &48.97 &12.90 &48.41 &15.08 &47.84 \\
        NoC (z=2) &1.53 &6.80 &6.45 &41.06 &4.38 &38.85 &4.95 &36.54 &4.42 &39.68 &3.81 &36.54 \\
        NoC (z=3) &1.99 &8.13 &11.42 &47.31 &8.86 &45.80 &10.36 &46.92 &8.72 &46.18 &8.26 &44.76 \\
        NoC (z=4) &5.36 &11.78 &27.88 &57.27 &24.06 &55.41 &23.72 &53.38 &23.65 &55.31 &25.61 &55.91 \\
        NoC (z=5) &9.19 &15.01 &42.27 &62.89 &38.59 &61.07 &35.13 &59.54 &35.57 &60.37 &50.73 &62.86 \\
        NoC (z=6) &11.79 &17.18 &\underline{49.28} &65.58 &44.00 &63.33 &39.53 &62.06 &41.25 &62.84 &56.00 &64.61 \\
        NoC (z=7) &12.11 &18.34 &49.18 &\textbf{66.65} &45.09 &\textbf{64.40} &39.25 &\textbf{63.02} &42.41 &\textbf{64.18} &57.87 &\textbf{65.21} \\
        NoC (z=8) &\underline{12.23} &\underline{18.43} &48.81 &\underline{66.31} &\underline{46.21} &\underline{64.10} &40.23 &\underline{62.84} &\underline{43.50} &\underline{63.77} &\underline{59.15} &64.11 \\
        
        \midrule\hline
        \rowcolor[gray]{0.85}\multicolumn{13}{l}{\textit{\textbf{Alignment Level Computation using ViT-B/32}}} \\ \hline
        NoC (z=1) &9.42 &13.79 &33.68 &54.00 &29.52 &50.99 &33.43 &51.32 &28.52 &50.82 &29.98 &51.23 \\
        NoC (z=2) &1.83 &7.26 &7.68 &43.43 &5.63 &41.52 &6.37 &43.09 &5.36 &42.00 &5.98 &40.16 \\
        NoC (z=3) &2.31 &8.73 &13.24 &49.10 &10.54 &47.36 &12.15 &48.49 &10.09 &47.62 &10.88 &46.51 \\
        NoC (z=4) &6.01 &12.56 &30.81 &59.03 &25.34 &56.72 &25.06 &56.10 &23.90 &56.44 &30.19 &57.41 \\
        NoC (z=5) &9.34 &15.29 &43.10 &63.38 &39.19 &61.33 &34.74 &59.24 &36.22 &60.70 &51.89 &63.14 \\
        NoC (z=6) &11.70 &17.18 &49.01 &65.23 &45.29 &63.22 &\underline{41.19} &61.09 &42.29 &62.86 &57.81 &64.53 \\
        NoC (z=7) &\textbf{12.32} &\textbf{18.67} &\textbf{49.68} &66.26 &\textbf{47.19} &63.86 &\textbf{42.07} &62.14 &\textbf{44.23} &63.51 &\textbf{60.31} &\underline{65.01} \\
        NoC (z=8) &11.31 &16.51 &47.77 &63.67 &43.71 &61.43 &39.70 &60.45 &41.11 &60.87 &54.91 &62.79 \\
        \bottomrule
    \end{tabular}
    }
    \vspace{-0.1cm}
    \caption{
        \small Zero-shot caption generation performances on MSCOCO and nocaps trained on CC3M when replacing the pre-trained CLIP ViT-L/14 with CLIP ViT-B/32 for alignment level computation. We observe our model seems robust to the pre-trained model. Numbers in \textbf{bold} and \underline{underlined} indicate the best and second-best ones for each model, respectively.
    }
    \vspace{-0.3cm}
    \label{table:zero_caption_vitb}
\end{table*}

\begin{table*}[!t]
    \centering
    \scalebox{0.92}{
        \begin{tabular}{l|rrr|rrr}
            \toprule
            \multirow{2}{*}{Models} &\multicolumn{3}{c|}{MSCOCO} &\multicolumn{3}{c}{Flickr30k} \\
            &R@1 &R@5 &R@10 &R@1 &R@5 &R@10 \\
    
            \midrule\hline
            \rowcolor[gray]{0.85}\multicolumn{7}{l}{\textit{\textbf{Alignment Level Computation using ViT-L/14}}} \\ \hline
            NoC (z=1) & 5.12 & 14.90 & 21.68 & 14.30 & 33.70 & 43.60 \\
            NoC (z=2) & 1.08 & 3.22 & 4.98 & 2.90 & 9.30 & 14.40 \\
            NoC (z=3) & 2.80 & 8.10 & 12.12 & 7.50 & 21.90 & 29.90 \\
            NoC (z=4) & 11.72 & 28.84 & 39.04 & 27.40 & 57.00 & 66.20 \\
            NoC (z=5) & 25.24 & 52.10 & 63.78 & 45.80 & 79.20 & 89.30 \\
            NoC (z=6) & 38.18 & 65.60 & 76.72 & 62.60 & 91.00 & 95.70 \\
            NoC (z=7) &\textbf{44.16} &\textbf{71.18} &\textbf{81.16} &\textbf{69.20} &\textbf{93.90} &\underline{97.20} \\
            NoC (z=8) & \underline{43.02} & \underline{70.74} & \underline{80.34} & \underline{67.90} & \underline{92.90} & \textbf{97.30} \\
            
            \midrule\hline
            \rowcolor[gray]{0.85}\multicolumn{7}{l}{\textit{\textbf{Alignment Level Computation using ViT-B/32}}} \\ \hline
            NoC (z=1) &9.26 &23.08 &31.12 &20.40 &45.80 &56.70 \\
            NoC (z=2) &1.38 &5.06 &7.82 &4.60 &14.10 &19.40 \\
            NoC (z=3) &3.92 &11.66 &17.70 &11.50 &29.00 &39.90 \\
            NoC (z=4) &16.22 &36.72 &47.88 &33.00 &64.40 &74.50 \\
            NoC (z=5) &28.66 &55.76 &66.82 &49.70 &82.30 &90.40 \\
            NoC (z=6) &\underline{36.40} &\underline{64.76} &\underline{75.92} &\underline{60.10} &\underline{91.00} &\underline{96.00} \\
            NoC (z=7) &\textbf{43.20} &\textbf{70.38} &\textbf{80.50} &\textbf{67.30} &\textbf{91.20} &\textbf{96.30} \\
            NoC (z=8) &29.50 &56.42 &68.64 &56.50 &84.50 &91.10 \\
            \bottomrule
        \end{tabular}
        \label{table:self_retrieval_cc3m_vitb}
    }
    \vspace{-0.1cm}
    \caption{
        \small Self-retrieval performance on MSCOCO and Flickr30k trained on CC3M when replacing the pre-trained CLIP ViT-L/14 with CLIP ViT-B/32 for alignment level computation. 
        Numbers in \textbf{bold} and \underline{underlined} indicate the best and second-best ones for each model, respectively.
    }
    \label{table:zero_retrieval_vitb}
\end{table*}

